\documentclass{article}
\usepackage[dvipsnames,table]{xcolor}
\usepackage[final]{corl_2026} 
\usepackage{algorithm}
\usepackage{algpseudocode}
\usepackage{amsfonts}
\usepackage{amsmath}
\usepackage{amsthm}
\usepackage{array}
\usepackage{booktabs}
\usepackage[font=footnotesize]{caption}
\usepackage{enumitem}
\usepackage{graphicx}
\usepackage[utf8]{inputenc}
\usepackage{multirow}
\usepackage{stfloats}
\usepackage[caption=false,font=normalsize,labelfont=sf,textfont=sf]{subfig}
\usepackage{svg}
\usepackage{textcomp}
\usepackage{url}
\usepackage{tabularx}
\usepackage{verbatim}

\usepackage{wrapfig}
\newcommand{\best}[1]{\textbf{\underline{#1}}}
\newcommand{\second}[1]{\textbf{#1}}
\newcommand{\third}[1]{\underline{#1}}

\newtheorem{proposition}{Proposition}

\title{SentryVLA: Uncertainty-Guided Action Selection and State Monitoring for Vision-Language-Action Robotic Control}

\author{
Zongyao Lyu$^{1,}$\thanks{Equal contribution.
$^{\dagger}$ Corresponding author: william.beksi@utdallas.edu.},
\hspace{0.5em}
Lingling Chen$^{2,*}$,
\hspace{0.3em}
William J. Beksi$^{2,\dagger}$\\[0.3em]
{\small
$^{1}$The University of Texas at Arlington
\hspace{2em}
$^{2}$The University of Texas at Dallas
}
}

\begin{document}
\nolinenumbers
\maketitle

\vspace{-8mm}
\begin{center}
\includegraphics[width=1\linewidth]{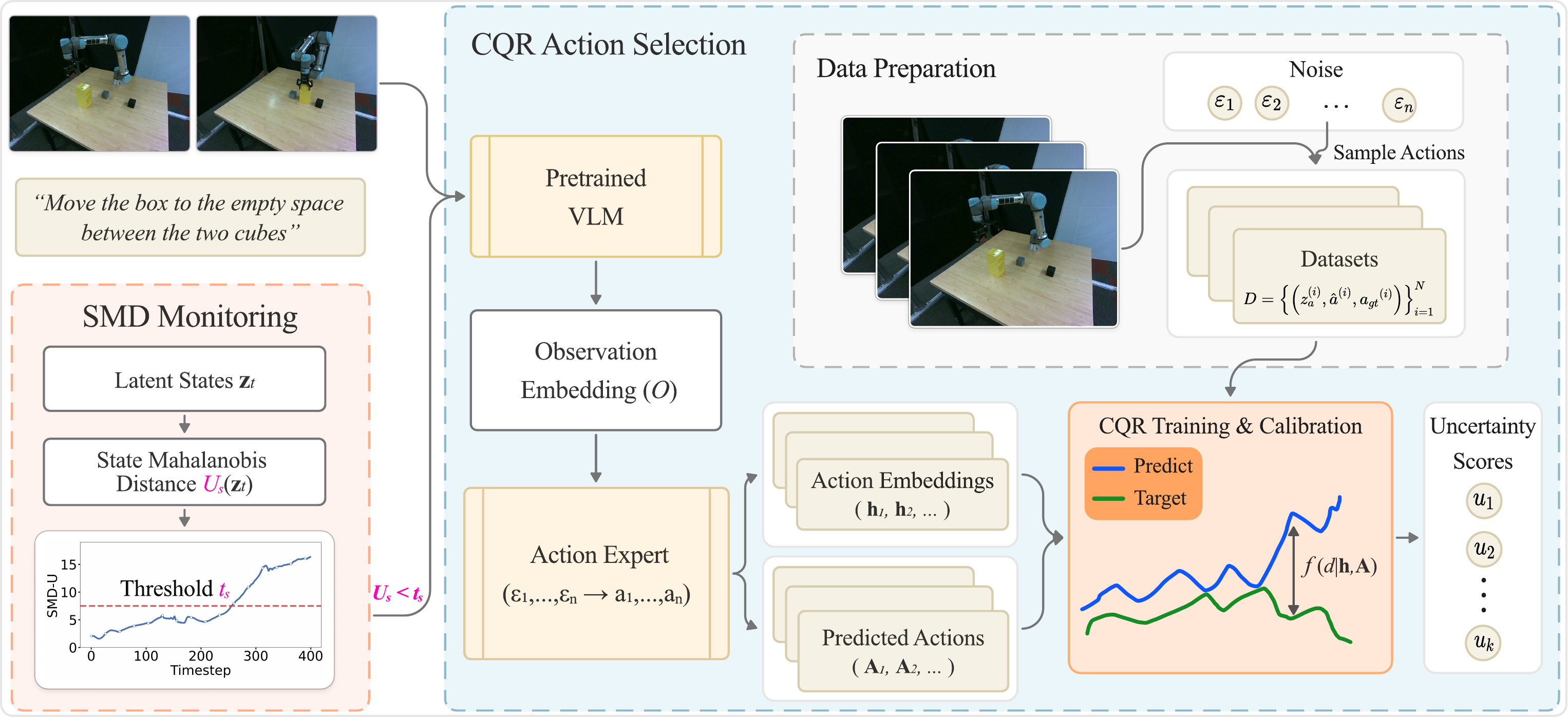}
\captionof{figure}{The SentryVLA framework consists of two main components: (i)
SMD Monitoring computes a Mahalanobis-distance anomaly score over
contextualized latent execution states to detect abnormal execution contexts;
(ii) CQR Action Selection ranks multiple noise-conditioned action candidates by
their learned action-deviation scores and selects the lowest-scoring candidate.
Together, these components enable uncertainty-guided action selection and
state-aware execution monitoring for VLA-based robotic control.}
\label{fig:overview}
\end{center}

\begin{abstract}
Vision-language-action (VLA) models have emerged as generalist robotic
controllers that map visual observations and language instructions to
continuous actions. However, pretrained VLAs typically provide limited
information about the reliability of their generated actions, making deployment
challenging when execution failures must be anticipated. We address this
problem by introducing SentryVLA, a lightweight runtime wrapper for frozen VLA
policies. At the action level, SentryVLA uses conformalized quantile regression
to estimate a calibrated upper bound on the deviation between sampled VLA
actions and expert-like behavior under the calibration distribution. The
predicted deviation is then used to rank multiple stochastic candidates for
action selection. At the execution-monitoring level, SentryVLA applies a
Mahalanobis-distance-based anomaly score to the contextualized latent execution
state extracted before final action prediction, providing a complementary
signal for detecting abnormal execution states during closed-loop control. The
framework does not require retraining or architectural modification of the
underlying VLA policy. Experiments in simulation and on a physical robot show
that SentryVLA improves action selection and provides useful runtime warning
signals for abnormal execution states. These findings demonstrate the practical
value of post-hoc reliability scoring for improving the runtime behavior of
pretrained VLA policies. Project webpage:
\href{https://robotic-vision-lab.github.io/sentryvla/}{https://robotic-vision-lab.github.io/sentryvla/}.
\end{abstract}


\section{Introduction}
\label{sec:introduction}
Recent progress in robot foundation models has shifted manipulation from
task-specific controllers toward generalist policies conditioned on images and
language~\citep{zhong2025survey}. Vision-language-action (VLA) models combine
visual-language pretraining with robot demonstrations to map multimodal
observations and instructions into executable actions. This generality expands
the scope of robot policy learning, but also exposes a reliability gap: the
scale, multimodal structure, and opaque action-generation process of VLAs make
runtime failures difficult to anticipate.

In deployment, a VLA-controlled robot may encounter lighting changes,
occlusions, clutter, ambiguous instructions, calibration errors, or execution
states outside the training distribution~\citep{liu2021towards,ganai2025real}.
Yet, pretrained VLAs typically output actions without a calibrated or directly
usable reliability signal. Common evaluations emphasize task success, language
grounding, or aggregate rollout performance, but do not indicate whether a
specific sampled action is risky or whether the current execution state is
drifting from nominal behavior. Without such signals, failures may only become
apparent after the robot has already entered an unsafe trajectory.

Uncertainty quantification and runtime monitoring offer tools for this problem.
Bayesian approximations, Monte Carlo dropout, deep ensembles, and conformal
prediction have been widely studied for predictive reliability in machine
learning~\citep{lyu2020probabilistic,rahaman2021uncertainty}. Robot learning
has used uncertainty for data aggregation, planning, teleoperation,
exploration, and risk-aware control. Nonetheless, VLA control is harder to
monitor because failures can arise jointly from perception, language grounding,
robot state evolution, and stochastic action generation.

Recent VLA reliability work addresses parts of this pipeline. Some methods use
planner confidence, token-level uncertainty, or calibration signals to request
help or intervene. Others evaluate action instability, train auxiliary
reliability models, classify upcoming failures, or use verifier-based test-time
sampling. A complementary direction monitors observations or internal
representations for visual interference, out-of-distribution (OOD) states, or
runtime failures. These studies show that VLA reliability can be estimated at
runtime, but they often focus on a single signal, require additional failure
labels or verifier models, or use uncertainty mainly as a diagnostic rather
than for action selection.

We introduce SentryVLA, a lightweight runtime wrapper for VLA policies.
SentryVLA targets two deployment questions: which sampled action should be
executed, and whether the current execution state appears abnormal. At the
action level, it uses conformalized quantile regression (CQR) to predict
deviation from expert-like behavior and to calibrate an upper bound on that
deviation under the calibration distribution. The predicted deviation is used
to rank multiple stochastic candidates. At the execution-monitoring level,
SentryVLA applies a Mahalanobis-distance-based anomaly score to the
contextualized latent execution state extracted before final action prediction,
providing a complementary signal for abnormal closed-loop states.

The two modules address different failure modes. The CQR scorer ranks
stochastic action candidates, while the latent-state monitor checks whether the
current execution context remains close to the nominal demonstration
distribution. This separation avoids conflating action-sample variability with
state abnormality and lets SentryVLA add reliability scoring around a
pretrained VLA without retraining or modifying the base policy
(Fig.~\ref{fig:overview}).

Our contributions are as follows: (i) quantile-based action-deviation scoring
for stochastic candidate selection with conformal calibration of
candidate-level deviation bounds; (ii) latent-state execution monitoring using
a Mahalanobis-distance score over the contextualized state representation
before final action prediction; (iii) integration of these components into a
frozen-policy runtime wrapper for uncertainty-guided action selection and
state-aware intervention; and (iv) simulation and physical-robot evaluation
showing improved stochastic action selection, useful abnormal-state warning
signals, and stronger success-failure discrimination when the two modules are
combined for reliability evaluation. We restrict conformal calibration claims
to the calibration distribution and treat OOD and hardware results as empirical
reliability tests rather than formal distribution-shift guarantees.

\section{Related Work}
\label{sec:related_work}
\textbf{Generalist robot policies.} Early generalist agents and
language-conditioned robot systems established two routes toward broad robot
behavior: shared transformer policies across modalities, as in
Gato~\citep{reed2022a}, and language-model planning over low-level skills, as
in SayCan~\citep{ichter2022do,chowdhery2023palm}. These systems showed that
large pretrained models can support robot decision-making, but they either
offered limited manipulation capability or depended on predefined skill
libraries. More recent embodied foundation models move toward integrated
perception-language-action control. PaLM-E~\citep{driess2023palm} incorporates
embodied observations into language-model reasoning, while
RT-2~\cite{brohan2023rt}, OpenVLA~\cite{kim2024openvla}, and
$\pi_0$~\cite{black2025pi_0} use visual-language pretraining and robot
demonstrations to produce robot actions across broad manipulation settings.
These models substantially improve generality, yet their outputs are usually
evaluated by aggregate success rates rather than calibrated runtime reliability
signals~\citep{firoozi2025foundation,kawaharazuka2025vision}.

\textbf{Uncertainty and calibration for robot learning.} Predictive uncertainty
has been studied through Bayesian neural networks, Monte Carlo dropout,
ensembles, conformal prediction, and related
methods~\citep{shafer2008tutorial,neal2012bayesian,gal2016dropout,lakshminarayanan2017simple,angelopoulos2021gentle}.
Bayesian and ensemble-based techniques estimate uncertainty through stochastic
inference or model diversity, while conformal prediction provides
distribution-free calibration of prediction sets or error bounds under its
calibration assumptions. These tools have been applied to vision, object
detection, and language modeling to identify unreliable predictions, OOD inputs,
or poorly calibrated confidence
estimates~\citep{miller2018dropout,lyu2021uncertainty,shelmanov2021certain,ulmer2022exploring,campos2024conformal,patel2025conformal}.
In robotics, uncertainty has been used for imitation-learning data aggregation,
uncertainty-aware planning, conformalized teleoperation, conformal
decision-making, and risk-sensitive
control~\citep{cui2019uncertainty,gershman2019uncertainty,sun2023conformal,lekeufack2024conformal,stachowicz2024racer,zhao2024conformalized}.
A further challenge in multimodal robotic systems is to distinguish uncertainty
arising from perception and language grounding from uncertainty in downstream
action generation, especially under input modality
shifts~\citep{dutta2023estimating,bhatt2025know}. SentryVLA follows this
operational use of uncertainty as a runtime decision signal, but focuses on a
lightweight post-hoc wrapper for pretrained VLA execution. 

\textbf{VLA reliability and failure detection.} Contemporary work has begun to
quantify when robots based on VLA or vision-language models (VLMs) should ask
for help, intervene, or distrust their outputs. Planner-level and token-level
methods use calibrated language-model confidence, entropy, perplexity, or
related scores as intervention
signals~\citep{ren2023robots,karli2025ask,zollo2025confidence}. Action-level
methods examine whether the generated action sequence itself is reliable, for
example, by evaluating action-sequence instability, training reliability models
on successful trajectories, predicting failures from partial trajectories, or
scaling test-time sampling with verifier
models~\citep{he2024rediffuser,gu2025safe,kwok2025robomonkey,valle2025evaluating}.
Observation- and representation-level methods instead monitor the inputs or
internal features that precede action generation to identify visual
interference, OOD states, or runtime failures, including VLM-based failure
explanation, visual observation intervention, density estimation over successful
embeddings, and joint input-output uncertainty for runtime
prediction~\citep{duan2025aha,hancock2025run,romer2025fiper,xu2025can}. These
methods demonstrate the value of runtime reliability estimation, but many
depend on failure supervision or verifier modules, and most use uncertainty to
assess generated actions rather than directly guide execution.  Recent
VLA-specific methods such as INSIGHT~\cite{karli2025insight} and
UPS~\cite{yuan2026act} also use model-internal uncertainty signals for help
triggering or policy steering. Compared with these lines of work, SentryVLA
couples two lightweight runtime signals in a single wrapper: conformalized
action-deviation scoring to rank stochastic action candidates, and Mahalanobis
monitoring of contextualized latent execution states to flag abnormal contexts
during closed-loop execution.

\section{Method}
\label{sec:method}
\subsection{Problem Formulation and Framework Overview}
\label{subsec:framework_overview}
We aim to equip a pretrained VLA policy with runtime reliability assessment
during real-world deployment. Specifically, we consider a frozen VLA policy
operating under receding-horizon control. At each decision step $t$, the policy
receives an observation
\begin{equation}
    \mathbf{o}_t = (\mathbf{I}_t, \ell, \mathbf{q}_t),
\end{equation}
where $\mathbf{I}_t$ denotes one or more RGB images, $\ell$ is a language
instruction, and $\mathbf{q}_t \in \mathbb{R}^{d_q}$ is the robot proprioceptive
state. The policy outputs either a single action or a short-horizon action chunk
\begin{equation}
    \mathbf{A}_t =
    [\boldsymbol{a}_t, \boldsymbol{a}_{t+1}, \ldots,
    \boldsymbol{a}_{t+H-1}],
    \qquad
    \boldsymbol{a}_t \in \mathbb{R}^{d_a}.
\end{equation}
In our single-arm manipulation setting, $\boldsymbol{a}_t$ is a continuous
end-effector action that includes a gripper command. The controller executes the
first action, or a short prefix of the action chunk, before receiving a new
observation and querying the policy again. The parameters of the base VLA remain
frozen throughout training, calibration, and deployment.

For a stochastic generative VLA, different noise samples $\epsilon_k$ may
produce multiple candidate action chunks from the same observation
$\mathbf{o}_t$:
\begin{equation}
    \mathbf{A}^{(k)}_t
    = \pi_\theta(\mathbf{o}_t; \epsilon_k),
    \qquad
    \epsilon_k \sim p(\epsilon),
    \qquad
    k = 1, \ldots, K.
\end{equation}
Although these candidates share the same visual, language, and robot-state
context, they may differ in execution quality. Moreover, the current execution
context may have shifted beyond the policy's nominal operating regime even
when the generated actions appear plausible. Reliable runtime decision-making
therefore requires both assessing uncertainty among candidate actions and
detecting abnormal execution contexts.

As shown in Fig.~\ref{fig:overview}, SentryVLA addresses these two issues using
two complementary modules. CQR Action Selection (CAS) scores the sampled action
candidates and selects the candidate with the lowest predicted deviation risk.
State Mahalanobis Distance (SMD) Monitoring evaluates the policy's latent
execution state
\begin{equation}
    \mathbf{z}_t = \phi_o(\mathbf{o}_t),
\end{equation}
where $\mathbf{z}_t$ is a policy-internal representation from the
action-prediction pathway. For $\pi_0$, we use the action-head hidden state at
the first denoising step after multimodal conditioning. Backbone-specific
extraction details are given in Appendix B.1. CAS addresses which action should
be selected under the current context, whereas SMD Monitoring determines
whether that context remains within the nominal operating regime. Together,
they form a post-hoc runtime reliability framework that does not require
retraining the base VLA. The following subsections describe the two modules and
their combined runtime procedure.

\subsection{CQR Action Selection (CAS)}
\label{subsec:action_scoring}
\textbf{Offline action-scoring dataset and training target.}
CAS learns a calibrated score for action-level uncertainty induced by
stochastic VLA sampling. We construct an offline action-scoring dataset by
querying the policy on demonstration or calibration observations and sampling
candidate actions. Each training sample stores
\begin{equation}
    \left(\mathbf{h}^{(k)}_{a,t}, \mathbf{A}^{(k)}_t, \mathbf{A}^{\mathrm{E}}_t\right),
\end{equation}
where $\mathbf{A}^{\mathrm{E}}_t$ is the expert action chunk for 
the same observation. The candidate-specific feature $\mathbf{h}^{(k)}_{a,t}$
is extracted from the action-generation pathway corresponding to the sampled
noise realization, and is paired with the generated action chunk
$\mathbf{A}^{(k)}_t$.  The action-deviation target is 
\begin{equation}
    d^{(k)}_t = \Delta\!\left(\mathbf{A}^{(k)}_t, \mathbf{A}^{\mathrm{E}}_t\right),
    \label{eq:action-deviation_target}
\end{equation}
where $\Delta(\cdot,\cdot)$ is an action-space deviation metric. In our
implementation, $\Delta$ is the Euclidean deviation over the continuous action
dimensions used for control. This target turns action-deviation estimation into
a supervised prediction problem. We use deviation from the paired expert action
as a ranking proxy for candidate selection, assuming that the local
demonstrations provide a meaningful reference for valid behavior. We examine
this assumption empirically in Appendix~\ref{app:cqr_score_design}.

\textbf{Quantile regression model.} Let
\begin{equation}
    \mathbf{x}^{(k)}_t = g\!\left(\mathbf{h}^{(k)}_{a,t}, \mathbf{A}^{(k)}_t\right)
\end{equation}
denote the input to the action-scoring model, implemented as the concatenation
of the candidate's action feature and predicted action representation. We train
a quantile regressor $f_\tau$ to estimate an upper conditional quantile of the
deviation target $d^{(k)}_t$:
\begin{equation}
    \hat{d}^{(\tau,k)}_t = f_\tau\!\left(\mathbf{x}^{(k)}_t\right).
\end{equation}
The model is optimized with the pinball loss
\begin{equation}
    \mathcal{L}_\tau(u) =
    \begin{cases}
        \tau u, & u \geq 0, \\
        (\tau - 1)u, & u < 0,
    \end{cases}
    \qquad
    u = d^{(k)}_t - \hat{d}^{(\tau,k)}_t .
\end{equation}
A high quantile level $\tau$ focuses the model on upper-tail action deviations,
which are most relevant for screening risky candidate actions.

\textbf{Conformal calibration.}
Quantile regression provides an initial action-deviation estimate, yet may be
miscalibrated. We therefore apply CQR~\citep{romano2019conformalized}. On a
held-out calibration set, we compute residual scores for each calibration
sample $j$:
\begin{equation}
    s_j = d_j - f_\tau(\mathbf{x}_j),
\end{equation}
and define the conformal correction offset
\begin{equation}
    q_\alpha = \operatorname{Quantile}_{1-\alpha}\!\left(\{s_j\}_{j=1}^{n_{\mathrm{cal}}}\right),
    \label{eq:conformal_calibration}
\end{equation}
using the finite-sample empirical quantile convention for the desired
miscoverage rate $\alpha$. The calibrated action-deviation score is
\begin{equation}
    U_a(\mathbf{x}) = f_\tau(\mathbf{x}) + q_\alpha.
    \label{eq:action-deviation_score}
\end{equation}
This score is a calibrated upper bound on expert-action deviation under the
calibration distribution. The global correction $q_\alpha$ places the model
prediction on a calibrated scale, while $f_\tau(\mathbf{x})$ provides the
candidate-dependent part of the score. Since $q_\alpha$ is shared across
candidates for a fixed calibration setting, it does not change their
within-observation ranking. Consequently, CAS ranks candidates according to
$f_\tau(\mathbf{x})$, while conformalization calibrates the candidate-level
deviation bound.

\textbf{Uncertainty-guided action selection.}
At runtime, the policy samples $K$ candidate chunks for the current
observation, where $K$ is the total number of candidates and
$k\in\{1,\ldots,K\}$ indexes an individual candidate. For each candidate, CAS
computes
\begin{equation}
    U^{(k)}_{a,t} = U_a\!\left(g\!\left(\mathbf{h}^{(k)}_{a,t}, \mathbf{A}^{(k)}_t\right)\right).
\end{equation}
The executed candidate is selected as the one with the smallest calibrated
action-deviation score,
\begin{equation}
    k^\ast = \arg\min_{k \in \{1,\ldots,K\}} U^{(k)}_{a,t},
    \qquad
    \mathbf{A}^{\ast}_t = \mathbf{A}^{(k^\ast)}_t .
    \label{eq:candidate_selection}
\end{equation}
After selecting $\mathbf{A}_t^\ast$, the controller executes its first action,
or a short prefix, and then re-queries the policy with a new observation. In
this way, stochastic action generation is converted into an uncertainty-guided
decision process. The base policy proposes multiple candidate actions and CAS
selects the one with the lowest predicted action-deviation score to execute.

\begin{proposition}[Marginal coverage]
\label{prop:coverage}
Let $f_\tau$ be fixed after training, and let
$\{(\mathbf{x}_j,d_j)\}_{j=1}^{n_{\mathrm{cal}}}$ be held-out calibration
frame-candidate samples, with $d_j$ defined by
Eq.~(\ref{eq:action-deviation_target}). Assume that, conditional on the
training data and training randomness, the calibration samples and a test
sample $(\mathbf{x},d)$ are exchangeable. We define
$s_j=d_j-f_\tau(\mathbf{x}_j)$ and
$r_\alpha=\lceil(n_{\mathrm{cal}}+1)(1-\alpha)\rceil$, and let $q_\alpha$ be
the $r_\alpha$-th smallest calibration residual with $q_\alpha=+\infty$ if
$r_\alpha>n_{\mathrm{cal}}$. Then, the calibrated bound in
Eq.~(\ref{eq:action-deviation_score}) satisfies
\begin{equation}
    \Pr\!\left(d\le U_a(\mathbf{x})\right)\ge 1-\alpha.
\end{equation}
The guarantee is marginal for an individual sampled candidate, and the scope
and proof are provided in Appendix~\ref{app:calibration_scope}.
\end{proposition}

\subsection{SMD Monitoring over Latent Execution States}
\label{subsec:latent_state_monitoring}
The action-selection module ranks candidates sampled under the current context,
but it does not evaluate whether the context itself is nominal. SMD Monitoring
addresses this complementary issue by scoring the VLA's contextualized latent
execution state $\mathbf{z}_t$ defined in Sec.~\ref{subsec:framework_overview}.
Compared with the raw proprioception state $\mathbf{q}_t$, this latent state
also reflects the current visual-language context; compared with final action
outputs, it remains upstream of action realization. Thus, SMD monitors the
execution context represented internally by the policy rather than a low-level
robot-state vector or an already-generated action.

\textbf{Nominal latent-state distribution.}
We collect latent execution states from nominal demonstrations or successful
calibration rollouts, $\mathcal{Z}_{\mathrm{nom}} =
\{\mathbf{z}_i\}_{i=1}^{N}$.  These states define the nominal latent execution
manifold. We fit a Gaussian reference model using the empirical mean and
regularized covariance:
\vspace{-1.5mm}
\begin{equation}
    \mu = \frac{1}{N}\sum_{i=1}^{N} \mathbf{z}_i,
    \qquad
    \Sigma_\lambda = \frac{1}{N-1}\sum_{i=1}^{N} (\mathbf{z}_i - \mu)(\mathbf{z}_i - \mu)^\top + \lambda I,
\label{eq:latent_mean_covariance}
\end{equation}
where $\lambda I$ is a regularization term used for numerical stability in
high-dimensional latent spaces. Appendix Table~\ref{tab:smd_density_comparison}
compares the Gaussian reference model with a Gaussian mixture model (GMM) and a
k-nearest-neighbors (k-NN) on LIBERO-10, supporting its use for SMD.

\textbf{State anomaly score.}
At runtime, the latent-state anomaly score is the Mahalanobis
distance~\citep{mahalanobis1936generalised} 
\begin{equation}
    U_s(\mathbf{z}_t) = \sqrt{(\mathbf{z}_t - \mu)^\top \Sigma_\lambda^{-1}(\mathbf{z}_t - \mu)}.
\end{equation}
Larger values indicate that the current latent execution state is farther from
the nominal distribution. We denote the anomaly-detection threshold by $t_s$.
With only nominal validation data provided, $t_s$ can be set as a high
empirical quantile of scores from successful executions. If labeled success and
failure validation rollouts are available, $t_s$ may instead be selected by
sweeping thresholds and optimizing a detection criterion such as Youden's
statistic. In either case, the monitor is a distance-based anomaly detector
whose output measures deviation from the nominal latent execution manifold. The
specific threshold selection used in our experiments is detailed in
Appendix~\ref{app:smd_threshold_sensitivity_analysis}.

At runtime, SMD computes $U_s(\mathbf{z}_t)$ before executing the next selected
action. If $U_s(\mathbf{z}_t) > t_s$, the current context is flagged as
abnormal and execution can be paused. The resulting state-monitoring signal
operates at the VLA policy frequency and supports runtime intervention. It
complements low-level robot safety controllers, which operate closer to the
actuator level.

\subsection{Combined Runtime Procedure}
\label{subsec:combined_runtime_procedure}
At runtime, SentryVLA combines state monitoring and action selection in a
single receding-horizon procedure. Offline, the action quantile model is
trained on sampled VLA actions with expert-reference deviations, the conformal
correction is calibrated on a held-out set, and the nominal latent-state
distribution is estimated for Mahalanobis monitoring. During deployment, each
policy step first computes the latent-state anomaly score and pauses execution
if the current context exceeds the state threshold. Otherwise, the policy
samples multiple candidate action chunks, SentryVLA scores them using the
calibrated action-deviation model, and the controller executes the candidate
with the lowest predicted action-deviation score before receiving the next
observation.  In summary, the state score acts as an execution-context gate,
while the action score ranks stochastic candidates for uncertainty-guided
control.  Algorithms~\ref{alg:cas} and~\ref{alg:smd} in the appendix present
module-level pseudocode for the two procedures.

\section{Experiments}
\label{sec:experiments}
\subsection{Experimental Setup}
\label{subsec:experimental_setup}
\textbf{VLA policies.} We evaluate SentryVLA on two representative VLA control
regimes: stochastic generative policies and deterministic action-head policies.
The stochastic policy is $\pi_0$, whose flow-matching action decoder generates
noise-conditioned action chunks and therefore supports CAS. The deterministic
policy is OpenVLA-OFT, featuring a continuous action head that produces a
single action chunk for a given observation. Hence, we evaluate both CAS and
SMD Monitoring on $\pi_0$, while OpenVLA-OFT is used only for SMD score
evaluation because it does not expose multiple stochastic candidates for CAS.

\textbf{Execution protocol and environments.} All policies are evaluated using
a first-action receding-horizon protocol. At each step, the policy predicts a
short-horizon action chunk, the controller executes only the first action, and
the policy is then re-queried with a new observation. In simulation, we use the
LIBERO-Object benchmark, which contains 10 object-manipulation tasks with
official evaluation protocols and 50 expert demonstrations per task. We use the
official fine-tuned $\pi_0$ and OpenVLA-OFT checkpoints. For hardware
evaluation, we fine-tune $\pi_0$ on 100 trajectories collected from six spatial
manipulation tasks and evaluate the same runtime wrapper on the physical robot. 

\textbf{Conformal calibration.} We set the quantile-regression level to
$\tau=0.9$ and the miscoverage rate to $\alpha=0.1$, corresponding to a target
marginal coverage level of $1-\alpha=0.9$. Data are split by trajectory,
keeping all frames and their sampled candidates from each trajectory in the
same split.  After training, we compute the conformal correction $q_\alpha$
according to Eq.~\ref{eq:conformal_calibration}, using all $K=10$ candidates
sampled at each frame in the held-out calibration set. Additional setup details
are given in Appendix~\ref{app:exp_setup}. The inference latency across
different candidate counts is reported in Table~\ref{tab:runtime_cost} in
Appendix~\ref{app:runtime_cost}.

\subsection{Evaluation Metrics and Baselines}
\label{subsec:evaluation_metrics}
For action selection, we report task success rate and compare CAS with three
baselines: the default policy, random selection among $K$ sampled candidates,
and mean action averaging over $K$ sampled candidates. For uncertainty and
runtime reliability assessment, we follow EU-VLA~\citep{valle2025evaluating}
and SAFE~\citep{gu2025safe}. We report Spearman correlation between reliability
score and task outcome, Vargha-Delaney's $A_{12}$~\citep{vargha2000critique},
Cohen's $d=2|A_{12}-0.5|$, and AUC for failure detection. Additionally, we
present accuracy as a thresholded success-versus-failure classification metric.
We compare against seven uncertainty metrics defined in the EU-VLA benchmark:
token probability (TB-TP), prediction confidence score (TB-PCS), entropy
(TB-E), DeepGini (TB-D), action position instability (A-PI), action velocity
instability (A-VI), and action acceleration instability (A-AI). Metric
definitions and baseline details are presented in
Appendix~\ref{app:metrics_and_baselines}.  For $\pi_0$, we also report CAS+SMD,
a weighted sum of the normalized CAS and SMD scores.  A weight-sensitivity
analysis is provided in Appendix~\ref{app:lambda_ablation}.

\section{Results}
\label{sec:results}
Before evaluating failure detection and task success, we verify that CQR
calibration brings empirical candidate-level coverage close to the target
levels $1-\alpha$ on held-out LIBERO-Object demonstrations
(Appendix~\ref{app:empirical_coverage}, Table~\ref{tab:cqr_coverage}).

\subsection{Runtime Reliability and Failure Detection}
\label{subsec:uncertainty_failure_evaluation}
\begin{table*}
\centering
\resizebox{\textwidth}{!}{%
\begin{tabular}{l|l|ccccc|ccccc}
\toprule
\multirow{2}{*}{\textbf{Category}} & \multirow{2}{*}{\textbf{Method}}
& \multicolumn{5}{c|}{\textbf{OpenVLA-OFT (LIBERO)}}
& \multicolumn{5}{c}{\textbf{$\pi_0$ (LIBERO)}}\\
& & $|\rho|$ ↑ & $A_{12}$ ↓ & $d$ ↑ & AUC ↑ & Acc ↑
& $|\rho|$ ↑ & $A_{12}$ ↓ & $d$ ↑ & AUC ↑ & Acc ↑\\
\midrule
\multirow{4}{*}{\textbf{Token-Based}~\citep{valle2025evaluating}}
& TB-TP
    & 0.231 & 0.168 & 0.664 & 0.832 & 0.926
    & 0.011 & 0.506 & 0.013 & 0.494 & 0.560\\
& TB-PCS
    & 0.218 & 0.186 & 0.629 & 0.814 & 0.926
    & 0.009 & 0.505 & 0.011 & 0.495 & 0.560\\
& TB-E
    & 0.192 & 0.224 & 0.552 & 0.776 & 0.924
    & 0.270 & 0.343 & 0.315 & 0.657 & 0.597\\
& TB-D
    & 0.187 & 0.231 & 0.538 & 0.769 & 0.922
    & 0.233 & 0.364 & 0.272 & 0.636 & 0.587\\
\midrule
\multirow{3}{*}{\textbf{Action-Based}~\citep{valle2025evaluating}}
& A-PI
    & 0.096 & 0.638 & 0.276 & 0.362 & 0.910
    & 0.643 & 0.125 & 0.750 & 0.875 & 0.727\\
& A-VI
    & 0.004 & 0.495 & 0.010 & 0.505 & 0.912
    & 0.650 & 0.121 & 0.757 & 0.879 & 0.740\\
& A-AI
    & 0.006 & 0.491 & 0.017 & 0.509 & 0.912
    & \third{0.659} & \third{0.116} & 0.768 & \third{0.884} & 0.743\\
\midrule
\multirow{3}{*}{\textbf{SentryVLA (Ours)}}
& \textbf{CAS}
    & -- & -- & -- & -- & --
    & \third{0.659} & \third{0.116} & \third{0.769} & \third{0.884} & \third{0.777}\\
& \textbf{SMD}
    & \second{0.252} & \second{0.137} & \second{0.725} & \second{0.863} & \best{0.934}
    & \second{0.723} & \second{0.078} & \second{0.843} & \second{0.922} & \second{0.825}\\
& \textbf{CAS + SMD}
    & -- & -- & -- & -- & --
    & \best{0.750} & \best{0.063} & \best{0.874} & \best{0.937} & \best{0.853}\\
\bottomrule
\end{tabular}%
}
\captionsetup{font=footnotesize,skip=2pt}
\caption{Evaluation metrics for OpenVLA-OFT and $\pi_0$ on LIBERO-Object.
Arrows indicate whether higher or lower values are better. CAS-based action
selection is evaluated only on stochastic $\pi_0$, while SMD applies to both
policies. For $\pi_0$, CAS+SMD denotes a weighted reliability-evaluation score
with $\lambda=0.35$. The best result is shown in bold and underlined, the
second best in bold, and the third best underlined. Overall, the combined score
achieves the strongest reliability performance, followed by SMD and CAS.}
\label{tab:uncertainty_evaluation_metrics}
\end{table*}

We evaluate whether the proposed scores distinguish successful from failed
executions. On LIBERO-Object, we collect 600 trajectories per policy: 341
successful and 259 failed rollouts for $\pi_0$, and 574 successful and 26
failed rollouts for OpenVLA-OFT. Table~\ref{tab:uncertainty_evaluation_metrics}
reports the quantitative comparison. SMD gives the strongest single-source
reliability signal on both policies. On $\pi_0$, combining the CAS
action-deviation score with the SMD latent-state anomaly score yields the best
overall success-failure discrimination, supporting their use as complementary
runtime signals.

\begin{table*}
\centering
\small
\setlength{\tabcolsep}{5pt}
\begin{tabular}{l|cc|cc|cc|cc}
\toprule
\multirow{2}{*}{\textbf{Method}}
& \multicolumn{2}{c|}{\textbf{LIBERO-Object}}
& \multicolumn{2}{c|}{\textbf{LIBERO-Spatial}}
& \multicolumn{2}{c|}{\textbf{LIBERO-Goal}}
& \multicolumn{2}{c}{\textbf{LIBERO-10}} \\
& Seen & Unseen & Seen & Unseen
& Seen & Unseen & Seen & Unseen \\
\midrule
SAFE-LSTM
& 0.593 & 0.794 & 0.553 & 0.560
& 0.580 & 0.572 & 0.625 & 0.677 \\
SAFE-MLP
& 0.930 & 0.929 & 0.551 & 0.509
& 0.513 & \textbf{0.630} & \textbf{0.741} & 0.683 \\
FAIL-Detect
& 0.946 & 0.904 & 0.543 & 0.549
& 0.644 & 0.497 & 0.640 & 0.473 \\
\midrule
CAS (Ours)
& 0.892 & 0.799 & \textbf{0.567} & 0.557
& 0.531 & 0.480 & 0.567 & 0.638 \\
SMD (Ours)
& \textbf{0.982} & 0.965 & 0.555 & \textbf{0.581}
& 0.687 & 0.599 & 0.641 & 0.709 \\
CAS + SMD (Ours)
& \textbf{0.982} & \textbf{0.970}
& \textbf{0.567} & \textbf{0.581}
& \textbf{0.696} & 0.599
& 0.643 & \textbf{0.741} \\
\bottomrule
\end{tabular}
\caption{Failure-detection ROC-AUC for $\pi_0$ on four LIBERO suites. Each
suite reserves three tasks per split for unseen-task evaluation. Bold
indicates the highest AUC in each column.}
\label{tab:cross_suite_failure_detection}
\end{table*}

In addition, we evaluate failure detection on all four LIBERO suites using a
separately collected set of 500 rollouts per suite from the unmodified $\pi_0$
policy. The base policy achieves success rates of 56.8\% on LIBERO-Object,
74.0\% on LIBERO-Spatial, 81.8\% on LIBERO-Goal, and 43.4\% on LIBERO-10. This
evaluation uses task-disjoint splits, task-wise trajectory truncation, and mean
per-task ROC-AUC, so its LIBERO-Object values are not directly comparable to
Table~\ref{tab:uncertainty_evaluation_metrics}.
Table~\ref{tab:cross_suite_failure_detection} compares SAFE and FAIL-Detect
with our method on seen and unseen tasks. Success rates and evaluation details
are provided in Appendix~\ref{app:cross_suite_detection}.

The score comparisons and distributions in Appendix
Figs.~\ref{fig:uncertainty_quantification_method_comparison}
and~\ref{fig:uncertainty_distribution} further illustrate these results.
Representative trajectories in Appendix Fig.~\ref{fig:smd_failure_detection}
show SMD remaining below the threshold during successful execution and crossing
it before failure in a failed rollout. The maximum-SMD distributions in
Appendix Fig.~\ref{fig:smd_distribution} reveal that successful rollouts are
similar to expert demonstrations, whereas failed rollouts exhibit a more
pronounced upper tail. Additional anomaly-detection statistics and threshold
sensitivity are reported in Appendix~\ref{app:sim_smd_detection}
and~\ref{app:smd_threshold_sensitivity_analysis}.

\subsection{Action Decision-Making Performance}
\label{subsec:action_decision-making_performance}
\setlength{\columnsep}{6pt} 
\begin{wraptable}{r}{0.35\columnwidth}
\vspace{-0.8em}
\centering
\captionsetup{font=footnotesize,skip=2pt}
\footnotesize
\setlength{\tabcolsep}{5pt}
\renewcommand{\arraystretch}{0.92}
\captionbox{Action selection performance on LIBERO-Object.
\label{tab:average_success_rate}}{%
\begin{tabular*}{0.32\columnwidth}{@{}l@{\hspace{0.3em}}c@{}}
\toprule
\textbf{Method} & \textbf{Success Rate} \\
\midrule
Default $\pi_0$ & 0.56 \\
Random Selection & 0.57 \\ 
Mean Action & 0.66 \\ 
\textbf{CAS} & \textbf{0.73} \\
\bottomrule
\end{tabular*}
}
\vspace{-1em}
\end{wraptable}
Having evaluated cross-suite failure detection, we next evaluate whether CAS
improves the actions executed by the stochastic policy on LIBERO-Object.  We
first compare three CQR-derived action scores and find that Euclidean action
deviation is the most predictive score, outperforming prediction interval width
and cosine similarity (Appendix Table~\ref{tab:cqr-based_uncertainty_scores}).
We then run each LIBERO-Object task for 30 trials and compare the default
$\pi_0$ policy, random candidate selection, mean action averaging, and CAS.
Table~\ref{tab:average_success_rate} summarizes the average result, with the
full per-task breakdown in Appendix Table~\ref{tab:success_rate}. CAS improves
the average success rate from 0.56 to 0.73, outperforming both random candidate
selection and mean action averaging. The improvement shows that the gain does
not come merely from sampling multiple actions, but from using the learned
action-deviation model to rank the sampled candidates. The per-task results in
Appendix Table~\ref{tab:success_rate} show improvements in 9 out of 10 tasks. 

\subsection{Real-World Deployment}
\label{subsec:real-world_deployment_results}
On the UR5 robot, we evaluate whether the same two signals transfer beyond
simulation. For success-failure separation, we collect 20 successful and 20
failed trajectories for each of the six tasks, totaling 240 real-robot
rollouts. As reported in Appendix Table~\ref{tab:real-robot_auc_scores}, SMD
achieves an AUC of 0.912 and CAS achieves 0.759, outperforming all evaluated
token-level and action-instability baselines. Appendix
Fig.~\ref{fig:robot_demonstrations} shows representative rollouts for the six
real-robot tasks.

We further evaluate action-level reliability on two selected tasks under mild
real-world perturbations, including lighting changes, camera-view shifts, and
clutter. Across 720 real-robot rollouts, CAS consistently improves over the
default $\pi_0$ policy and is competitive with or better than mean action
averaging across the evaluated task-condition pairs (Appendix
Table~\ref{tab:ur5_success_rate}).  Finally, in workspace-edge trials, default
$\pi_0$ triggers protective stops in 20/20 trials, while SMD gating reduces
protective stops to 4/20 and proactively halts 16/20 trajectories before the
robot enters the failure region (Appendix
Table~\ref{tab:real_robot_failure_detection} and
Fig.~\ref{fig:proactive_vs_protective}).  These hardware results support the
same conclusion as simulation: CAS supports uncertainty-guided stochastic
action selection, while SMD provides a useful abnormal-context warning signal
during deployment.

\section{Limitations}
\label{sec:limitations}
SentryVLA is a post-hoc reliability wrapper whose guarantees depend on
calibration data and nominal demonstrations. CAS estimates calibrated
expert-action deviation; hence it may penalize valid actions that differ
significantly from the demonstrations. SMD detects deviations from nominal
latent execution states, but may miss failures near the demonstrated
latent-state manifold. Our hardware evaluation is limited to one UR5
manipulator setup, a small set of spatial tasks, and workspace-edge trials that
test only a controlled failure mode. Future work will evaluate broader tasks
and embodiments, adapt thresholds under sequential distribution shift, and add
richer multimodal representations or recovery policies.

\section{Conclusion}
\label{sec:conclusion}
We presented SentryVLA, a lightweight runtime wrapper for improving the
reliability of pretrained VLA policies without retraining or modifying the base
model. SentryVLA combines CAS, which uses CQR to select among stochastic action
candidates based on calibrated expert-deviation scores, with SMD Monitoring,
which detects abnormal execution contexts through Mahalanobis scoring over
contextualized latent execution states. Experiments in simulation and on a real
robot show that CAS improves stochastic action selection, SMD provides useful
abnormal-state warning signals, and their combination further improves
success-failure discrimination. These results support post-hoc reliability
scoring as a practical direction for safer deployment of generalist VLA
policies.

\section*{Acknowledgments}
\label{sec:acknowledgments}
This work was supported in part by the Naval Engineering Education Consortium
(NEEC) managed by the Naval Surface Warfare Center Dahlgren Division (NSWCDD)
under grant N00178-25-1-0036.

\bibliography{sentryvla_uncertainty-guided_action_selection_and_state_monitoring_for_vision-language-action_robotic_control}

\clearpage

\appendix
\label{sec:appendix}
\section{Additional Methodology Details}
\label{app:additional_methodology_details}
This appendix provides the method and evaluation details that support
Secs.~\ref{sec:method}--\ref{sec:results}. Appendix~\ref{app:cas_algorithm}
augments the CAS procedure from Sec.~\ref{subsec:action_scoring};
Appendix~\ref{app:smd_algorithm} expands the SMD Monitoring procedure from
Sec.~\ref{subsec:latent_state_monitoring}; Appendix~\ref{app:calibration_scope}
clarifies the calibration and reliability-evaluation scope.
Appendix~\ref{app:exp_setup}--\ref{app:runtime_cost} contain the experimental
details and CQR score-design comparison. Appendix~\ref{app:sim_results}
documents the full simulation results, Appendix~\ref{app:sensitivity_analysis}
provides the sensitivity analyses, and Appendix~\ref{app:real_robot} reports
the real-robot results.

\subsection{CQR Pipeline and Action-Selection Algorithm}
\label{app:cas_algorithm}
\begin{center}
\includegraphics[width=0.6\linewidth]{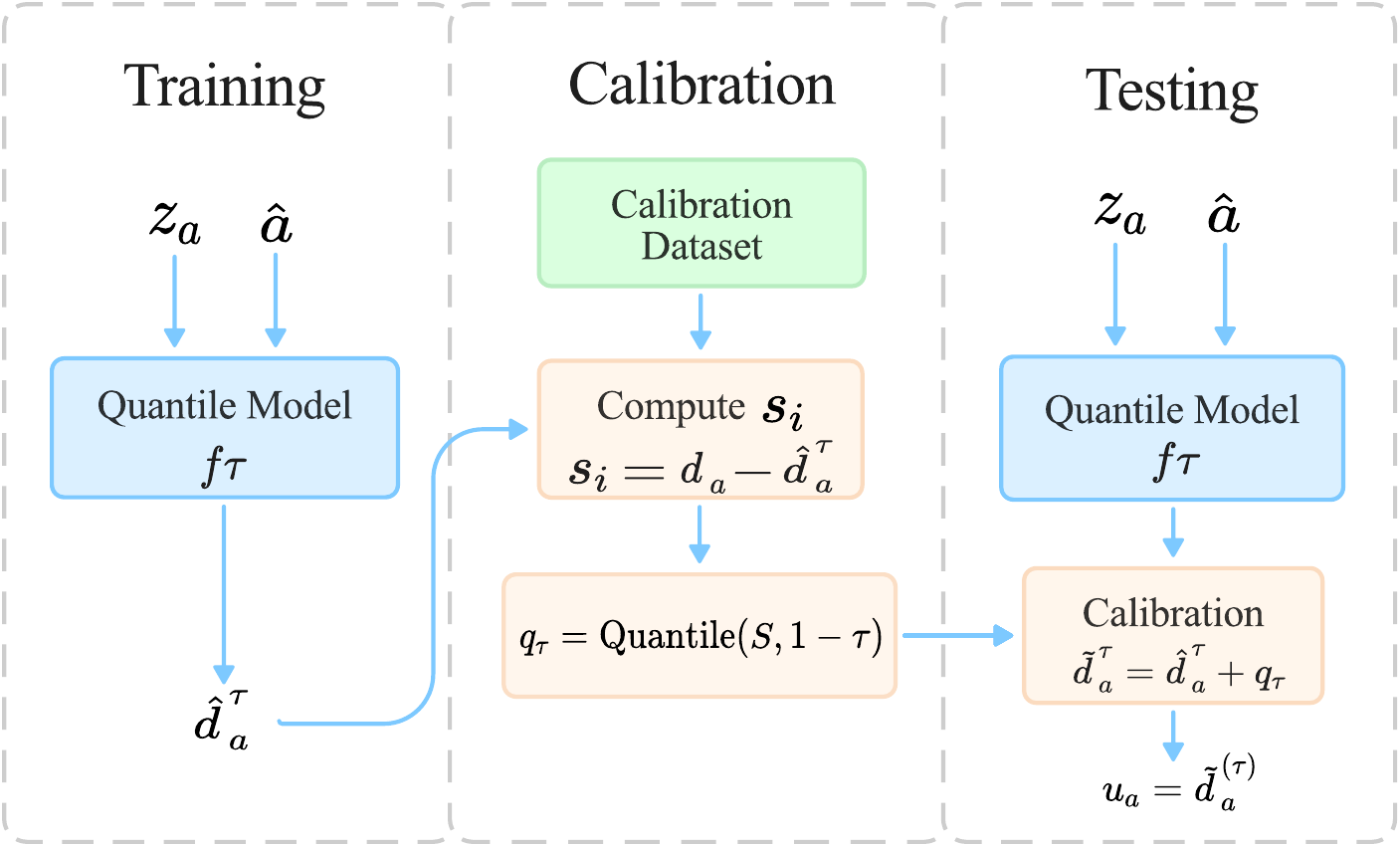}
\captionof{figure}{The CQR pipeline for calibrated action-deviation scoring.
Throughout training, a quantile regression model learns to predict an upper
quantile of the deviation between a sampled action and its expert-reference
action using offline data. During calibration, a held-out set is used to
compute the conformal correction offset. During testing, the offset is added to
the quantile prediction to score each sampled candidate, and the candidate with
the smallest calibrated action-deviation score is selected.}
\label{fig:cqr_pipeline}
\end{center}

Fig.~\ref{fig:cqr_pipeline} shows the training, calibration, and testing stages
of the CAS scoring model. The offline sampled-action dataset is constructed as
described in Sec.~\ref{subsec:action_scoring}: a frozen VLA is queried on
demonstration or calibration observations under sampled noise, and each sampled
action is paired with its action feature and expert-reference action.
Expert-reference actions are employed only for offline target construction and
conformal calibration. At runtime, CAS uses only the frozen VLA, the trained
quantile regressor, and the conformal offset.

The action-deviation target is computed over the continuous action dimensions
used for control. In our single-arm setting, the gripper command is binary; it
is treated as a separate control component and is excluded from the Euclidean
continuous-action deviation used for CAS. This target definition is kept fixed
across training, calibration, validation, and runtime scoring.

Algorithm~\ref{alg:cas} provides the runtime candidate-selection procedure. The
algorithm assumes that the quantile model and conformal offset have already
been fit offline. For each observation, the stochastic VLA generates $K$
noise-conditioned action chunks. CAS scores every candidate with the calibrated
action-deviation model and executes the candidate with the smallest score.

\begin{algorithm}[t]
\caption{CQR Action Selection (CAS)}
\label{alg:cas}
\begin{algorithmic}[1]
\Require Frozen VLA policy $\pi_\theta$, trained quantile regressor $f_\tau$,
conformal offset $q_\alpha$, current observation $\mathbf{o}_t$, number of
candidates $K$
\Ensure Selected action chunk $\mathbf{A}^{*}_t$
\State $\mathcal{A}_t \leftarrow \emptyset$, $\mathcal{H}_t \leftarrow \emptyset$, $\mathcal{U}_t \leftarrow \emptyset$
\For{$k=1$ to $K$}
    \State Sample $\epsilon_k \sim p(\epsilon)$
    \State $(\mathbf{A}^{(k)}_t,\mathbf{h}^{(k)}_{a,t}) \leftarrow \pi_\theta(\mathbf{o}_t;\epsilon_k)$
    \State $\mathcal{A}_t \leftarrow \mathcal{A}_t \cup \{\mathbf{A}^{(k)}_t\}$
    \State $\mathcal{H}_t \leftarrow \mathcal{H}_t \cup \{\mathbf{h}^{(k)}_{a,t}\}$
\EndFor
\For{$k=1$ to $K$}
    \State $\mathbf{x}^{(k)}_t \leftarrow g(\mathbf{h}^{(k)}_{a,t},\mathbf{A}^{(k)}_t)$
    \State $\hat d^{(\tau,k)}_t \leftarrow f_\tau(\mathbf{x}^{(k)}_t)$
    \State $U^{(k)}_{a,t} \leftarrow \hat d^{(\tau,k)}_t + q_\alpha$
    \State $\mathcal{U}_t \leftarrow \mathcal{U}_t \cup \{U^{(k)}_{a,t}\}$
\EndFor
\State $k^* \leftarrow \arg\min_{k\in\{1,\ldots,K\}} U^{(k)}_{a,t}$
\State $\mathbf{A}^{*}_t \leftarrow \mathbf{A}^{(k^*)}_t$
\State \Return $\mathbf{A}^{*}_t$
\end{algorithmic}
\end{algorithm}

\subsection{SMD Pipeline and Monitoring Algorithm}
\label{app:smd_algorithm}
SMD Monitoring utilizes the contextualized latent execution state
$\mathbf{z}_t$ before final action prediction, as defined in
Sec.~\ref{subsec:framework_overview}. This differs from the raw proprioceptive
state and from the final action output. The monitor fits a Gaussian reference
model to latent states extracted from nominal demonstrations and successful
calibration rollouts, and then scores a runtime state by its Mahalanobis
distance from this nominal latent-state distribution. This is a fixed
distance-based anomaly detector. 
 
Algorithm~\ref{alg:smd} gives the offline fitting and runtime evaluation
procedure for SMD monitoring. The covariance is regularized by the ridge term
$\lambda I$, matching Eq.~\ref{eq:latent_mean_covariance} in the main method.
The threshold $t_s$ is selected from validation rollouts as described in
Appendix \ref{app:smd_threshold_sensitivity_analysis}. At runtime, crossing
this threshold flags the current execution context as abnormal.

\begin{algorithm}[t]
\caption{SMD Monitoring over Latent Execution States}
\label{alg:smd}
\begin{algorithmic}[1]
\Require Nominal latent-state set $\mathcal{Z}_{\mathrm{nom}}=\{\mathbf{z}_i\}_{i=1}^{N}$, latent-state extractor $\phi_o$, current observation $\mathbf{o}_t$, regularization
coefficient $\lambda$, threshold $t_s$
\Ensure Nominal/abnormal decision for the current execution context
\Statex \textbf{Offline fitting}
\State $\boldsymbol{\mu} \leftarrow \frac{1}{N}\sum_{i=1}^{N}\mathbf{z}_i$
\State $\boldsymbol{\Sigma}_\lambda \leftarrow \frac{1}{N-1}\sum_{i=1}^{N}(\mathbf{z}_i-\boldsymbol{\mu})(\mathbf{z}_i-\boldsymbol{\mu})^\top + \lambda I$
\State Precompute $\boldsymbol{\Sigma}^{-1}_\lambda$
\State Select $t_s$ from validation rollouts
\Statex \textbf{Runtime evaluation}
\State $\mathbf{z}_t \leftarrow \phi_o(o_t)$
\State $U_s(\mathbf{z}_t) \leftarrow \sqrt{(\mathbf{z}_t-\boldsymbol{\mu})^\top\boldsymbol{\Sigma}^{-1}_\lambda(\mathbf{z}_t-\boldsymbol{\mu})}$
\If{$U_s(\mathbf{z}_t)>t_s$}
    \State \Return Abnormal
\Else
    \State \Return Nominal
\EndIf
\end{algorithmic}
\end{algorithm}

\subsection{Calibration and Reliability-Evaluation Scope}
\label{app:calibration_scope}
The CAS score is calibrated with respect to the offline calibration
distribution and the expert-action deviation target. Therefore, its conformal
interpretation applies to the calibration distribution under the standard
exchangeability assumption. This guarantee applies to an individual sampled
candidate; it does not imply coverage after CAS selection or simultaneous
coverage across a trajectory. In closed-loop rollouts, distribution-shift
settings, and physical-robot deployment, we use the score as an empirical
runtime reliability signal rather than claiming formal distribution-shift
coverage. SMD is a Mahalanobis-distance anomaly monitor with a
validation-selected threshold, not a conformal predictor. Its output measures
deviation from the nominal latent execution-state distribution and is used for
abnormal-context detection and intervention.

\paragraph{Proof of Proposition~\ref{prop:coverage}.} Let $n:=n_{\mathrm{cal}}$
and $s_{n+1}:=d-f_\tau(\mathbf{x})$. We condition on the training data and
training randomness so that $f_\tau$ is fixed. By assumption, the calibration
residuals $s_1,\ldots,s_n$ and the test residual $s_{n+1}$ are exchangeable.

Let $r_\alpha=\lceil(n+1)(1-\alpha)\rceil$. For $r_\alpha\le n$, attach an
independent $\mathrm{Unif}(0,1)$ variable to each residual and order the
resulting pairs lexicographically to break ties. Then, the rank $R$ of the test
pair among all $n+1$ residuals is uniform on $\{1,\ldots,n+1\}$. Since $R\le
r_\alpha \implies s_{n+1}\le q_\alpha$,
\begin{equation}
\begin{aligned}
    \Pr\!\left(d\le U_a(\mathbf{x})\right)
    &\ge \frac{r_\alpha}{n+1}
    &\ge 1-\alpha.
\end{aligned}
\end{equation}
This inequality holds conditionally on training and hence also marginally. If
$r_\alpha>n$, then $q_\alpha=+\infty$ and coverage is one.

\section{Additional Experiment Details}
\label{app:experiments_results}
\subsection{Experimental Setup}
\label{app:exp_setup}
\textbf{Policy regimes.} The experiments employ $\pi_0$ and OpenVLA-OFT to
cover stochastic and deterministic VLA control regimes. The $\pi_0$ policy uses
a flow-matching action decoder, thus repeated sampling with different noise
variables produces multiple candidate action chunks for the same observation.
This is the setting in which CAS is evaluated for action selection. OpenVLA-OFT
uses a deterministic continuous action head for a given observation. Therefore,
it is included to evaluate SMD and the EU-VLA reliability metrics under a
deterministic policy setting, but not for CAS-based stochastic candidate
selection.

\textbf{Latent-state extraction.} For $\pi_0$, SMD uses the action-head hidden
state at the first denoising step after multimodal conditioning. For
OpenVLA-OFT, SMD uses the hidden representation immediately before the action
head. In both cases, the representation is extracted from the frozen policy
before the final action prediction.

\textbf{Action execution strategy.} The $\pi_0$ model generates actions in
short-horizon chunks. For consistent evaluation, we adopt a receding-horizon
execution strategy whereby, at each timestep, only the first action of each
predicted chunk is executed before obtaining a new observation and generating
the next chunk. All action selection and state monitoring procedures operate
within this rolling execution framework.

\textbf{LIBERO simulation protocol.} The simulation experiments use the
LIBERO-Object suite with 10 object-manipulation tasks. Each task provides 50
expert demonstrations, and we utilize the official fine-tuned $\pi_0$ and
OpenVLA-OFT checkpoints. The reported stochastic action-selection experiments
sample $K=10$ candidate action chunks per observation for $\pi_0$ and execute
the candidate selected by the evaluated rule. Each LIBERO-Object task is
executed for 30 trials when measuring task success.

\textbf{Real-robot protocol.} 
The hardware experiments use a UR5 manipulator equipped with a Robotiq gripper
and two Intel RealSense D435 cameras. The cameras provide the visual
observation used by the policy. Robot execution is implemented in ROS 2 using
the Universal Robots driver and MoveIt Servo. The policy outputs a 7D Cartesian
delta action, consisting of three translational components, three rotational
components, and one gripper command. Policy inference is served remotely using
the OpenPI stack, while the robot-side client receives the predicted action
chunks and executes them through the ROS 2 control stack. We use
receding-horizon execution. At each policy query, only the first action of the
selected chunk is executed before acquiring a new observation and querying the
policy again. The deployed $\pi_0$ policy is fine-tuned on 100 real-world
trajectories collected across six manipulation tasks: (i) move the wipes bottle
near the red cup; (ii) move the red container near the blue bin; (iii) move the
tissue box farther from the orange cup; (iv) move the yellow box to the empty
space between the two cubes; (v) pick up the gray cube and place it on the
white plate; and (vi) pick up the red cup and place it in the black container.
The real-robot experiments evaluate three aspects: success-failure score
separation, action-selection robustness under mild scene perturbations, and SMD
gating in a workspace-edge failure setting.

\subsection{Evaluation Metrics and Baselines}
\label{app:metrics_and_baselines}
\subsubsection{Metrics}
To evaluate the proposed runtime signals, we adopt three complementary families
of metrics. The first two follow EU-VLA~\citep{valle2025evaluating}, which
assesses how uncertainty correlates with task performance and how well it
separates successful and failed executions. The third follows
SAFE~\citep{gu2025safe}, which evaluates reliability scores as binary failure
indicators.

\textbf{Uncertainty-performance correlation.} We measure the monotonic
relationship between runtime scores and task outcomes using Spearman rank
correlation. A strong negative correlation ($|\rho|\rightarrow 1$) indicates
that higher uncertainty or anomaly score is consistently associated with task
failure.

\textbf{Inter-group separability.} To quantify the discriminative power of a
score between successful and failed executions, we report Vargha-Delaney's
$A_{12}$~\cite{vargha2000critique}, which measures the probability that scores
from failed executions exceed those from successful ones, and Cohen's
$d=2|A_{12}-0.5|$, which captures the standardized difference between the two
groups.

\textbf{Failure detection reliability.} Following SAFE-style evaluation, we
treat a runtime score as a binary classifier over safe versus unsafe states and
compute AUC. A higher AUC indicates more reliable failure detection across
decision thresholds.

\paragraph{Cross-suite failure detection.}
\label{app:cross_suite_detection}
We evaluate the unmodified $\pi_0$ policy over 500 rollouts per LIBERO suite,
collected separately from the 600 LIBERO-Object rollouts used in
Table~\ref{tab:uncertainty_evaluation_metrics}. The numbers of successful
rollouts are 284 on Object (56.8\%), 370 on Spatial (74.0\%), 409 on Goal
(81.8\%), and 217 on LIBERO-10 (43.4\%). Seen/unseen tasks are defined relative
to detector fitting, rather than base-policy training. For each of five random
seeds, three tasks per suite are held out for unseen-task testing. Episodes
from each remaining task are split into 50\% training, 25\% calibration, and
25\% seen-task testing sets, with no episode overlap. All methods use the same
splits and task-wise trajectory truncation protocol, with each episode scored
by its maximum over time. We report mean per-task ROC-AUC for seen and unseen
tasks separately.

\subsubsection{Baselines}
\textbf{Action-selection baselines.} For action selection on stochastic
$\pi_0$, we compare CAS with: (i) default policy execution, which directly
executes the first sampled action; (ii) random selection, which randomly
selects one action from the $K$ sampled candidates; and (iii) mean action,
which takes the element-wise average of the $K$ sampled actions.

\textbf{Candidate-dispersion baseline.} We additionally evaluate label-free
uncertainty signals derived from the dispersion of the $K$ stochastic action
candidates. These signals use only the sampled policy outputs and do not
require expert-action targets.

\textbf{Reliability-score baselines.} For the reliability assessment, we
include seven uncertainty metrics from EU-VLA. Token probability (TB-TP)
averages the maximum token probabilities and converts them into an uncertainty
score, where lower token confidence produces higher uncertainty. The prediction
confidence score (TB-PCS) measures the margin between the largest and
second-largest token probabilities, where a smaller margin indicates greater
ambiguity between candidate tokens. Entropy (TB-E) computes the entropy of the
token probability distribution and is higher when the probability mass is
spread across many possible tokens. DeepGini (TB-D) quantifies the impurity of
the token distribution, with larger values indicating less concentrated token
predictions. Action position instability (A-PI) gauges abrupt changes in the
predicted action sequence through first-order action differences. Action
velocity instability (A-VI) measures oscillations in the decision process
through variation in action velocity. Action acceleration instability (A-AI)
assesses higher-order fluctuations in the action sequence through changes in
acceleration, capturing abrupt and less smooth action dynamics.

\textbf{Combined CAS+SMD score.} For $\pi_0$, where both CAS and SMD are
available, we also report a weighted reliability-evaluation score combining the
normalized action-level CAS score and latent-state SMD score. The full
definition and sensitivity analysis are detailed in
Appendix~\ref{app:lambda_ablation}. This combined score is used only for
reliability evaluation, not for CAS-based action selection.

\subsection{Empirical Coverage Evaluation}
\label{app:empirical_coverage}
We evaluate empirical coverage on 454 LIBERO-Object expert demonstrations,
split by trajectory into 225 training and 229 held-out trajectories. We train
three quantile regressors at $\tau\in\{0.95,0.90,0.80\}$ and use $K=10$
candidates sampled at each frame. Held-out trajectories are randomly divided
into approximately equal-sized calibration and test sets, keeping all frames
and candidates from each trajectory in the same split. For each regressor, we
set $\alpha=1-\tau$ and compute $q_\alpha$ via
Eq.~\ref{eq:conformal_calibration}.  We repeat the experiment with three random
training/held-out partitions. For each partition, we randomly divide the
held-out trajectories into calibration and test sets 200 times.

\begin{wraptable}{r}{0.5\linewidth}
\vspace{-0.2em}
\centering
\captionsetup{font=footnotesize,skip=2pt}
\footnotesize
\setlength{\tabcolsep}{3pt}
\renewcommand{\arraystretch}{0.95}
\caption{Candidate-level coverage before and after CQR calibration on
LIBERO-Object.}
\label{tab:cqr_coverage}
\begin{tabular}{@{}ccccc@{}}
\toprule
$\alpha$ & $\tau$ &
\shortstack{Uncalibrated\\coverage} &
\shortstack{Calibrated\\coverage} &
$q_\alpha$ \\
\midrule
0.05 & 0.95 & 0.944 & $0.950\pm0.004$ & $+0.052$ \\
0.10 & 0.90 & 0.894 & $0.900\pm0.006$ & $+0.020$ \\
0.20 & 0.80 & 0.788 & $0.800\pm0.007$ & $+0.023$ \\
\bottomrule
\end{tabular}
\vspace{-0.8em}
\end{wraptable}
Before calibration, we assess the quantile regressors by measuring the
proportion of test candidates satisfying $d\le f_\tau(\mathbf{x})$. As shown in
Table~\ref{tab:cqr_coverage}, empirical coverage is close to, but slightly
below, the nominal quantile levels $\tau$. After adding the conformal
correction $q_\alpha$, the proportion satisfying $d\le
f_\tau(\mathbf{x})+q_\alpha$ is closer to the target coverage $1-\alpha$ at all
three levels. This evaluation concerns individual candidates before CAS
selection at demonstration states.

\subsection{CQR Score Design}
\label{app:cqr_score_design}
\textbf{Data preparation and CQR model training.} To learn the action-deviation
score, we train a CQR model on data collected from the $\pi_0$ policy. We
generate a dataset $\mathcal{D}$ by running $\pi_0$ on the training tasks under
20 different random seeds, introducing varying noise into the action decoder.
Each timestep provides an offline training sample consisting of: (i)
action-level features $\mathbf{h}_{a,t}$, (ii) model-predicted action chunks
$\mathbf{A}_t$, and (iii) corresponding expert-reference actions
$\mathbf{A}^{\mathrm{E}}_t$ from the demonstration data. The CQR model takes as
input the concatenation of the action feature and the predicted action
representation, and is trained using the pinball loss to predict a high
quantile of the action-deviation target. The quantile level $\tau$ and
miscoverage rate $\alpha$ are provided in Sec.~\ref{subsec:experimental_setup}.

\textbf{Candidate action-deviation scores.} We compare three CQR-derived score
designs on validation rollouts. Let $\hat{\boldsymbol{a}}_t$ denote the
predicted continuous action vector at step $t$, and let
$\boldsymbol{a}^{\mathrm{E}}_t$ denote the corresponding expert-reference
action. The superscript $(6)$ denotes the six continuous end-effector
dimensions, excluding the gripper command. The symbols
$\hat{\boldsymbol{a}}^{\mathrm{low}}_t$ and
$\hat{\boldsymbol{a}}^{\mathrm{up}}_t$ denote lower and upper quantile
predictions for interval-width scoring. The expert action appears only in the
offline target used to train or evaluate the score design. At runtime, CAS uses
the trained quantile model and does not require expert actions.

\textbf{Prediction interval width (PIW-CQR).} This score measures the width of
the predicted action interval:
\begin{equation}
    U_t^{\mathrm{PIW}} =
        \max\left(
            \left\| \hat{\boldsymbol{a}}_t - \hat{\boldsymbol{a}}_t^{\mathrm{low}} \right\|_2,
            \left\| \hat{\boldsymbol{a}}_t^{\mathrm{up}} - \hat{\boldsymbol{a}}_t \right\|_2
        \right).
\end{equation}

\textbf{Cosine similarity score (COS-CQR).} This score measures directional
agreement between the predicted and expert-reference actions:
\begin{equation}
    c_t =
        \frac{\hat{\boldsymbol{a}}_t \cdot \boldsymbol{a}_t^{\mathrm{E}}}
        {\|\hat{\boldsymbol{a}}_t\|_2\,\|\boldsymbol{a}_t^{\mathrm{E}}\|_2},
        \qquad
        U_t^{\mathrm{COS}} = 1 - \hat{c}_t^{(0.9)}.
\end{equation}
Here, $\hat c_t^{(0.9)}$ is the predicted 0.9-quantile of the cosine similarity
score used for reliability ranking.

\textbf{Euclidean distance score (EUC-CQR).} This score directly measures
continuous action-space deviation:
\vspace{-3mm}
\begin{equation}
    U_t^{\mathrm{EUC}} =
        \left\|
        \hat{\boldsymbol{a}}_t^{(6)} - \boldsymbol{a}_t^{\mathrm{E},(6)}
        \right\|_2.
\end{equation}
\vspace{-3mm}

\begin{wraptable}{r}{0.5\linewidth}
\vspace{-0.8em}
\centering
\captionsetup{font=footnotesize,skip=2pt}
\footnotesize
\setlength{\tabcolsep}{4pt}
\renewcommand{\arraystretch}{0.95}
\caption{A comparison of CQR-derived action-deviation scores. EUC-CQR gives the
strongest correlation and separability and is the default CAS score.}
\label{tab:cqr-based_uncertainty_scores}
\begin{tabular}{@{}lccc@{}}
\toprule
\textbf{Method} &
\textbf{$\rho$ ↓} &
\textbf{$A_{12}$ ↓} &
\textbf{$d$ ↑} \\
\midrule
PIW-CQR & -0.742 & 0.067 & 0.865 \\
COS-CQR & -0.389 & 0.273 & 0.453 \\
\textbf{EUC-CQR} &
\textbf{-0.781} &
\textbf{0.045} &
\textbf{0.910} \\
\bottomrule
\end{tabular}
\vspace{-0.8em}
\end{wraptable}

Table~\ref{tab:cqr-based_uncertainty_scores} shows that the Euclidean action
deviation (EUC-CQR) has the strongest negative correlation with task success
and the largest effect size, providing the strongest validation signal among
the evaluated CQR-derived scores. Consequently, we adopt the Euclidean distance
score as the default action-deviation score for CAS in the main experiments.

\textbf{Expert-deviation assumption.} We further examine whether the Euclidean
action-deviation target is primarily driven by motion phase, motion speed, or
variation among valid expert behaviors. Using the same action-space deviation
metric as in Eq. (6), we analyze 20,000 expert frames. The ratio of free-space
to contact deviations is 1.02, while the mean deviations for fast and slow
motions are 0.179 and 0.178, respectively, indicating little dependence on
either factor. In contrast, the mean cross-episode expert-to-expert deviation
is 0.272, exceeding the policy-to-expert deviation of 0.158. These results
suggest that motion phase and speed do not substantially explain the deviation
score, while multimodal expert behavior remains an important limitation of
using deviation from a single reference action as the training target.

\subsection{Runtime Cost}
\label{app:runtime_cost}
\begin{wraptable}{r}{0.45\linewidth}
\vspace{-0.8em}
\centering
\captionsetup{font=footnotesize,skip=2pt}
\footnotesize
\setlength{\tabcolsep}{3pt}
\renewcommand{\arraystretch}{0.95}
\caption{Policy inference latency with different numbers of sampled
candidates.}
\label{tab:runtime_cost}
\begin{tabular}{@{}lrrrr@{}}
\toprule
Candidates ($K$) & 1 & 5 & 10 & 20 \\
\midrule
\shortstack[l]{Policy latency (ms)}
& 121.3 & 174.2 & 230.9 & 359.8 \\
\bottomrule
\end{tabular}
\vspace{-0.8em}
\end{wraptable}

We measure inference latency on an NVIDIA A40 GPU. The candidates share a
single visual-language prefix computation, while action-expert denoising is
batched across $K$ candidates. Table~\ref{tab:runtime_cost} reports policy
inference latency at different candidate counts. The shared prefix takes
approximately 67 ms. At $K=10$, policy inference takes 230.9 ms. SMD and CQR
scoring take 0.319 ms and 0.601 ms, respectively, for a combined overhead of
approximately 0.40\% relative to policy inference. On this benchmark, 230.9 ms
corresponds to approximately 4.3 policy inferences per second. Candidate
generation is therefore the main runtime cost.

\section{Additional Results}
\label{sec:additional_results}
\subsection{Full Simulation Results}
\label{app:sim_results}
\begin{figure*}[t]
\centering
\setlength{\abovecaptionskip}{2pt}
\includegraphics[width=1\linewidth]{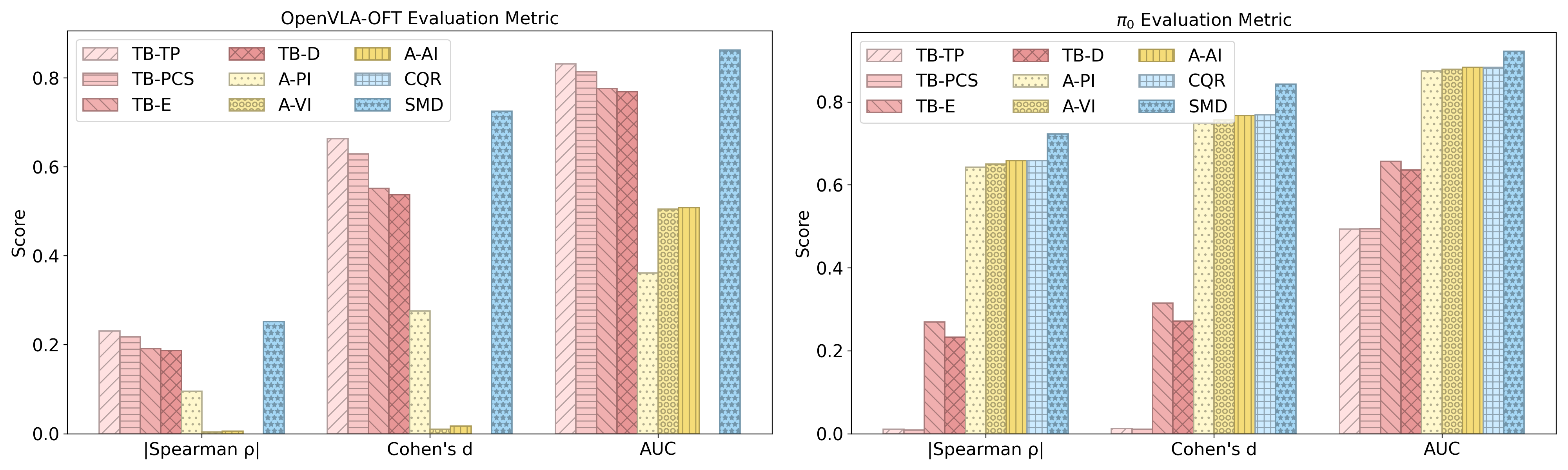}
\caption{A comparison of runtime reliability metrics on OpenVLA-OFT (left) and
$\pi_0$ (right) across Spearman $\rho$, Cohen's $d$, and AUC. The plot
summarizes the single-source scores. SMD provides the strongest single-source
signal, while CAS is evaluated on the stochastic $\pi_0$ setting where
noise-conditioned action candidates are available. The CAS+SMD combined score
for $\pi_0$ is reported in Table~\ref{tab:uncertainty_evaluation_metrics} and
analyzed in Appendix~\ref{app:lambda_ablation}.}
\label{fig:uncertainty_quantification_method_comparison}
\vspace{-4mm}
\end{figure*}

\begin{figure*}[!t]
\centering
\includegraphics[width=1\linewidth]{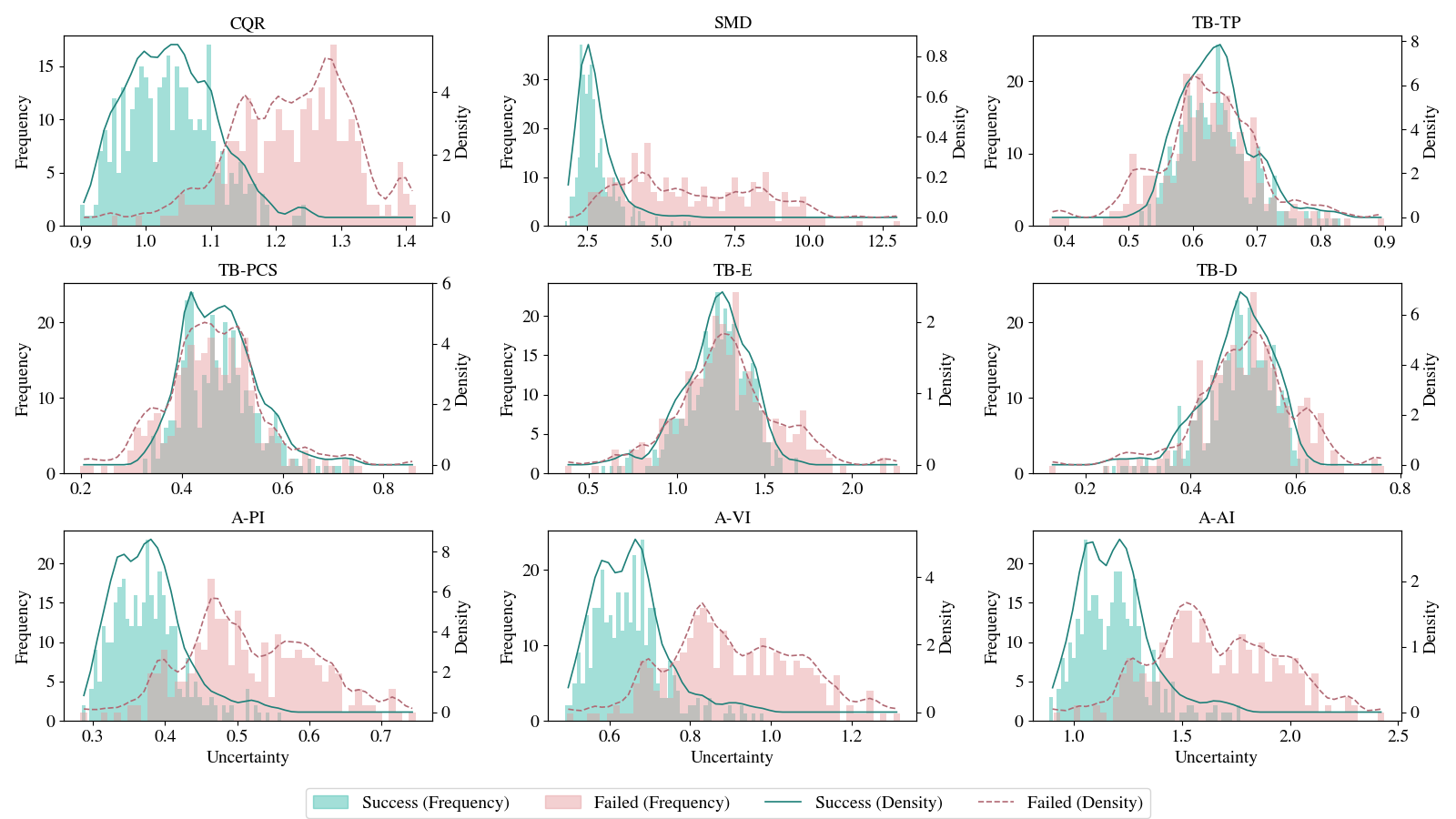}
\caption{The distribution of runtime reliability scores for successful and
failed executions across different reliability metrics. Each subplot shows how
a given method separates successes (cyan) from failures (pink) using frequency
histograms and kernel density estimates. A stronger separation indicates better
reliability. CAS and SMD produce a clearer separation than token-level
baselines.}
\label{fig:uncertainty_distribution}
\end{figure*}

We gather a test set of 600 execution trajectories on LIBERO-Object: 341
successful and 259 failed runs for $\pi_0$, and 574 successful and 26 failed
runs for OpenVLA-OFT. Table~\ref{tab:uncertainty_evaluation_metrics} in the
main paper gives the complete quantitative comparison.
Fig.~\ref{fig:uncertainty_quantification_method_comparison} provides the
corresponding visual summaries. Token-based metrics are weak for $\pi_0$, while
action-instability baselines are stronger because stochastic sample variation
is visible in the action sequence. SMD gives the strongest single-source score
on both policies. For $\pi_0$, CAS and SMD capture complementary action-level
and latent-state signals, and their weighted combination gives the strongest
overall reliability-evaluation result. Fig.~\ref{fig:uncertainty_distribution}
further shows that CAS and SMD produce a clearer separation between successful
and failed rollouts than token-level baselines, while action-instability
metrics provide moderate separation on $\pi_0$.

\begin{figure*}[t]
\centering
\includegraphics[width=1\linewidth]{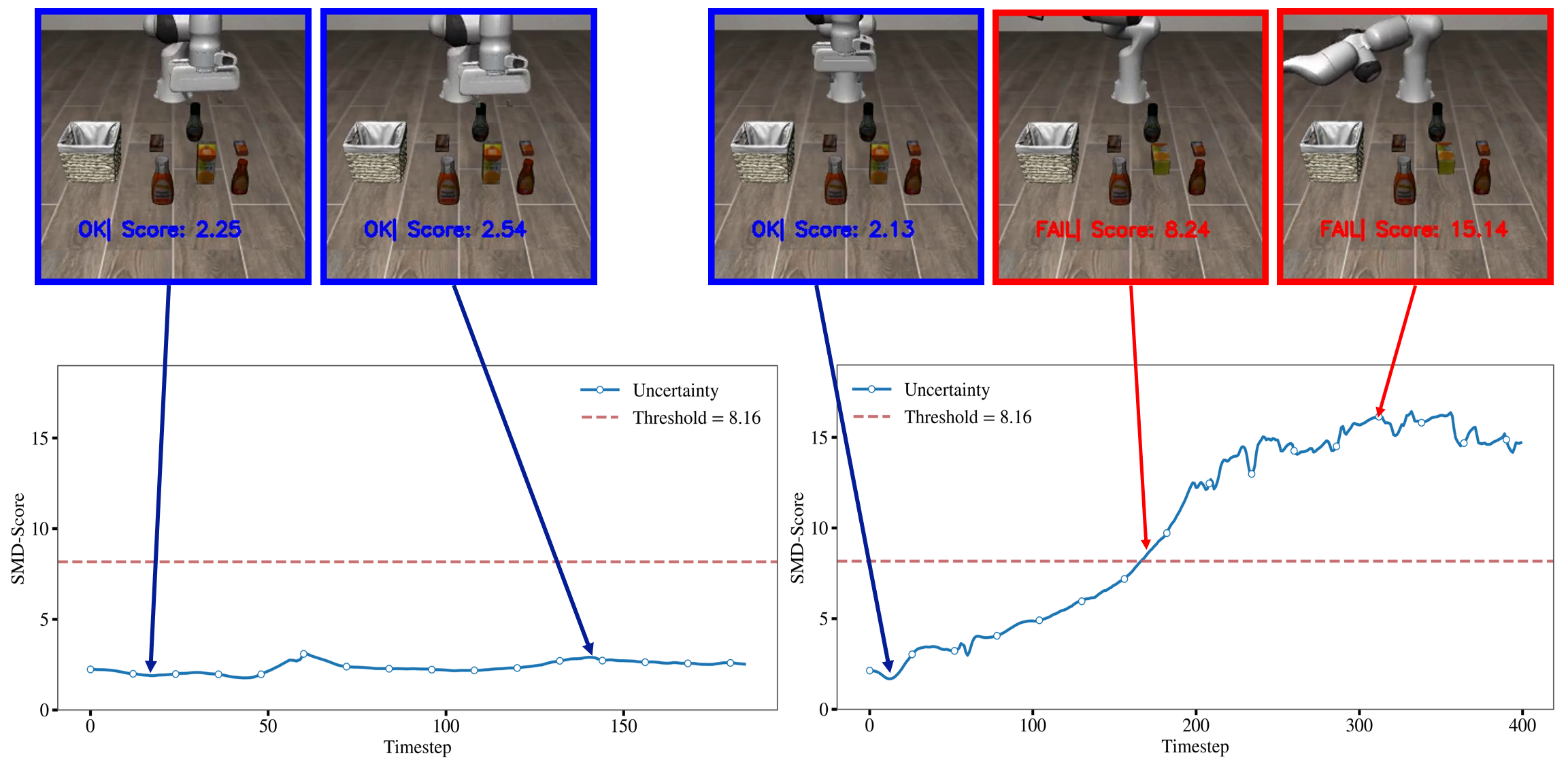}
\caption{Examples of SMD Monitoring in simulation. The left example shows a
successful execution where SMD stays below the threshold (red dashed line,
$t_s=8.16$), indicating that the latent execution state remains close to the
nominal distribution. The right example shows a failed execution where SMD
rises above the same threshold as the robot approaches a tipping-risk state.}
\label{fig:smd_failure_detection}
\end{figure*}

Beyond aggregate statistics, Fig.~\ref{fig:smd_failure_detection} illustrates
when SMD raises alarms during execution. Successful executions maintain low SMD
values throughout. By contrast, failed rollouts often show a rising SMD score
before the final failure event, providing an early warning signal before the
robot enters a visible failure state.

\subsubsection{Action Decision-Making Performance}
\label{app:sim_action_selection}
\begin{table}[t]
\centering
\begin{tabular}{lcccc}
\toprule
\textbf{Task} & \textbf{Default} & \textbf{Random} & \textbf{Mean} & \textbf{CAS} \\
\midrule
alphabet soup & 0.67 & 0.53 & 0.73 & \textbf{0.77} \\
cream cheese & 0.60 & 0.43  & \textbf{0.67} & 0.63 \\
salad dressing & 0.53 & 0.57  & 0.67 & \textbf{0.77} \\
BBQ sauce & 0.70 & 0.73 & 0.83 & \textbf{0.93} \\
ketchup & 0.90 & 0.93 & 0.97 & \textbf{1.00} \\
tomato sauce & 0.40 & 0.53 & 0.50 & \textbf{0.80} \\
butter & 0.60 & 0.73 & 0.77 & \textbf{0.87} \\
milk & 0.60 & 0.50 & \textbf{0.63} & 0.47\\
chocolate pudding & 0.57 & 0.67 & 0.67 & \textbf{0.83} \\
orange juice & 0.03 & 0.07 & 0.20 & \textbf{0.23} \\
\midrule
\textbf{Average} & 0.56 & 0.57 & 0.66 & \textbf{0.73} \\
\bottomrule
\end{tabular}
\captionsetup{skip=3pt}
\caption{The average success rate over 30 independent executions for each
LIBERO-Object task. Bold entries denote the best-performing method within each
row.}
\label{tab:success_rate}
\vspace{-0.8em}
\end{table}

Table~\ref{tab:success_rate} provides the per-task breakdown behind
Table~\ref{tab:average_success_rate} in the main paper and reports success
rates on all 10 LIBERO-Object tasks. Mean action averaging improves over
default execution, achieving an average success-rate increase of $+0.10$. CAS
provides a larger average improvement of $+0.17$ and outperforms both random
candidate selection and mean action averaging. 

\textbf{Candidate-dispersion results.} Candidate dispersion provides some
reliability information but is not consistent across suites. The evaluated
dispersion score obtains ROC-AUC 0.703, 0.511, 0.519, and 0.621 on
LIBERO-Object, LIBERO-Spatial, LIBERO-10, and LIBERO-Goal, respectively. On
LIBERO-10, a supervised probe over all 13 dispersion statistics reaches 0.656
AUC. These results suggest that stochastic sample dispersion alone is not a
reliable substitute for the action-deviation model across the evaluated suites.

\subsubsection{SMD Anomaly Detection Results}
\label{app:sim_smd_detection}
We further evaluate SMD as an anomaly gate over simulated rollouts. Before SMD
gating, the evaluation set contains 600 rollouts, including 341 successful and
259 failed executions, with 65 unsafe-state cases observed. As summarized in
Table~\ref{tab:sim_smd_detection}, applying the SMD threshold $t_s=8.16$ flags
159 rollouts while retaining 441 for continued execution, reduces the
failed-rollout count from 259 to 103, and removes the observed unsafe-state
cases. These results support the role of SMD as a latent-state anomaly monitor
complementary to CAS.

\begin{table}[t]
\centering
\begin{tabular}{lcccc}
\toprule
 & \textbf{Rollouts} & \textbf{Success / Failure} & \textbf{Flagged Rollouts} & \textbf{Unsafe Cases} \\
\midrule
Before SMD gate & 600 & 341 / 259 & 0 & 65 \\
After SMD gate ($U_s<t_s$) & 441 & 338 / 103 & 159 & 0 \\
\bottomrule
\end{tabular}
\captionsetup{skip=3pt}
\caption{SMD anomaly-gating statistics on LIBERO-Object. The SMD gate filters
high-anomaly rollouts and removes the observed unsafe-state cases while
retaining most successful rollouts.}
\label{tab:sim_smd_detection}
\end{table}

\begin{figure}[t]
\centering
\includegraphics[width=0.7\linewidth]{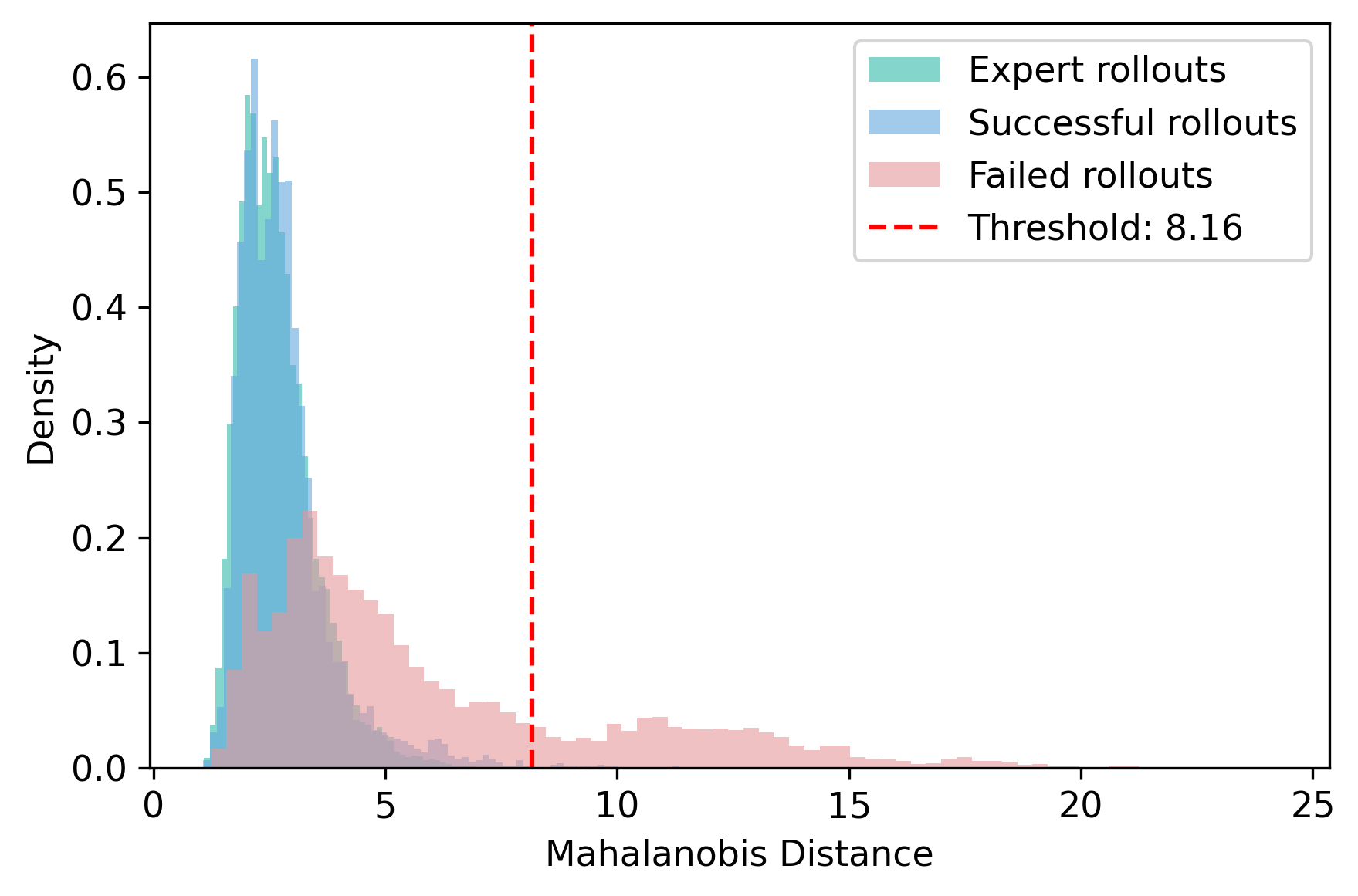}
\caption{The distributions of the maximum SMD per trajectory for expert
demonstrations, successful rollouts, and failed rollouts.}
\label{fig:smd_distribution}
\end{figure}

\paragraph{Latent-state anomaly scores.} We compare the maximum SMD per
trajectory across expert demonstrations, successful rollouts, and failed
rollouts. Fig.~\ref{fig:smd_distribution} shows that successful rollouts have a
score distribution similar to expert demonstrations, whereas failed rollouts
exhibit a more pronounced upper tail. The overlap between successful and failed
rollouts indicates that SMD does not distinguish all failures.

\paragraph{Anomaly-score comparison.} We contrast the Gaussian reference model
used by SMD with GMM and k-NN anomaly detectors on LIBERO-10, using the same
latent features and evaluation splits. Table~\ref{tab:smd_density_comparison}
shows that for the latent execution state used by SMD, the Gaussian model gives
the strongest held-out performance and generalizes more consistently than GMM
or k-NN. Raw 8D proprioception is substantially weaker on held-out tasks even
with the more flexible alternatives. These results support both the latent
execution representation and the lightweight Gaussian reference model in this
setting. We do not assume that the latent distribution is intrinsically
Gaussian.

\begin{table}[t]
\centering
\begin{tabular}{@{}lccc@{}}
\toprule
\textbf{Model} & \textbf{Train} & \textbf{Seen} & \textbf{Unseen} \\
\midrule
Proprioception (8D), Gaussian & 0.662 & 0.619 & 0.493\\
Proprioception (8D), GMM & 0.751 & 0.654 & 0.514\\
Proprioception (8D), k-NN & 0.849 & 0.700 & 0.543\\
\midrule
Latent execution state, Gaussian (SMD) & 0.798 & \textbf{0.709} & \textbf{0.641} \\
Latent execution state, GMM & \textbf{0.959} & 0.593 & 0.422 \\
Latent execution state, k-NN & 0.822 & 0.658 & 0.552 \\
\bottomrule
\end{tabular}
\captionsetup{skip=3pt}
\caption{Failure-detection ROC-AUC on LIBERO-10 for alternative SMD
representations and density models.}
\label{tab:smd_density_comparison}
\end{table}

\subsection{Sensitivity Analysis}
\label{app:sensitivity_analysis}
\subsubsection{SMD Threshold Sensitivity Analysis}
\label{app:smd_threshold_sensitivity_analysis}
\begin{figure}[t]
\centering
\includegraphics[width=0.65\linewidth]{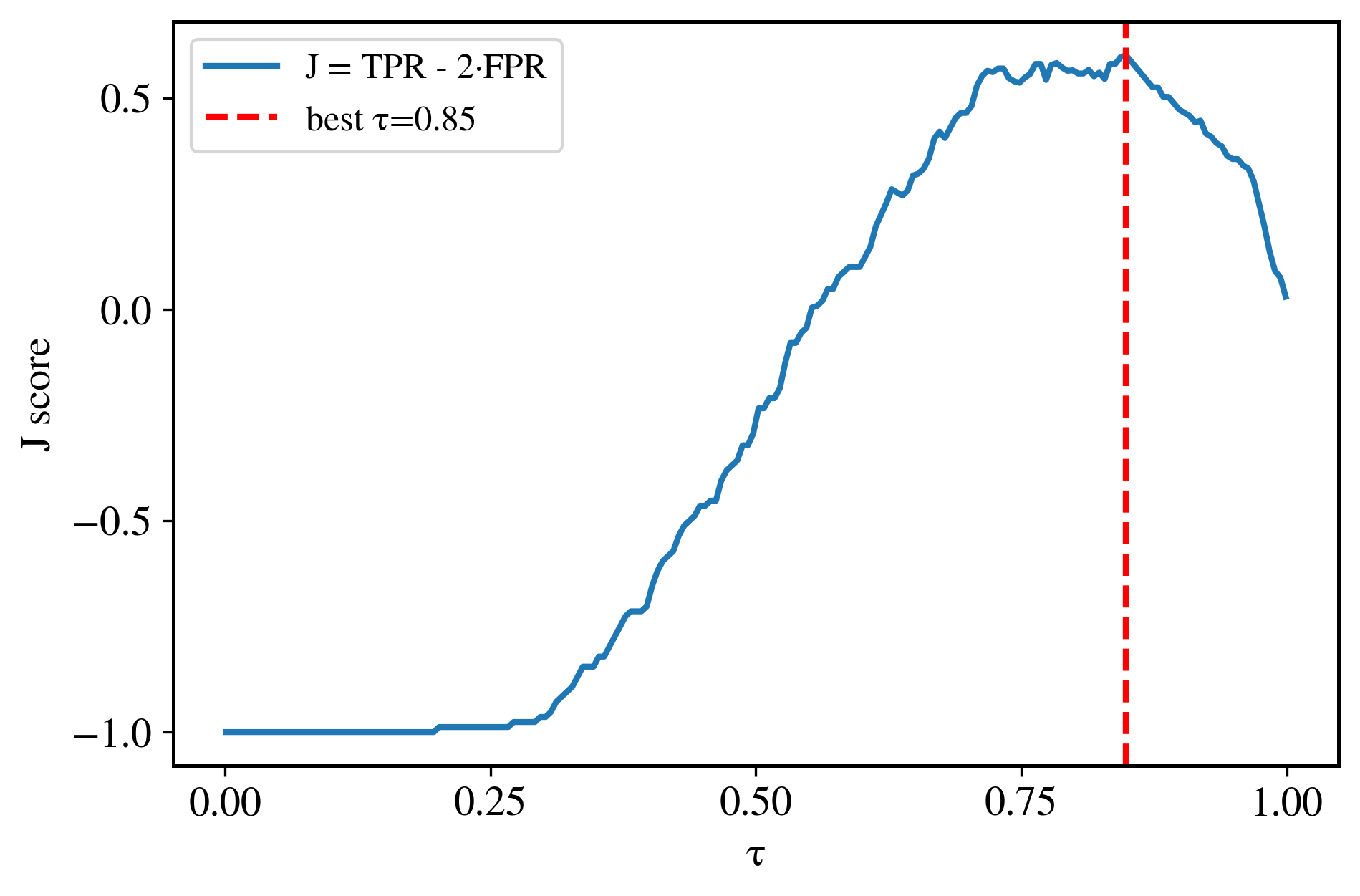}
\captionsetup{skip=-1pt}
\caption{The weighted $J_{\beta}$ score as a function of confidence level
$\tau$. The maximum is achieved at $\tau=0.85$ (red dashed line), which is
adopted as the confidence level for the SMD threshold.}
\label{fig:smd_j_score}
\end{figure}

To identify a threshold for triggering intervention based on SMD, we use a
weighted Youden's $J$-style criterion~\cite{fluss2005estimation}. For each
trajectory in the calibration dataset, we compute the maximum SMD score over
the trajectory and associate it with a binary success/failure label. We sweep
confidence levels $\tau$ and derive the corresponding threshold $t_s$ using
empirical quantiles of the SMD scores. At each $\tau$, we calculate
\begin{equation}
    J_{\beta} = \mathrm{TPR} - \beta\cdot \mathrm{FPR},
\end{equation}
where TPR is the true positive rate, FPR is the false positive rate, and
$\beta=2$ penalizes false positives more heavily. Fig.~\ref{fig:smd_j_score}
shows that the maximum is achieved at $\tau=0.85$, corresponding to threshold
$t_s=8.16$. This threshold identifies collapse-risk failure segments while
introducing limited disruption, reducing the overall task success rate by only
0.006 in the reported analysis.

\subsubsection{CAS + SMD Weight Sensitivity}
\label{app:lambda_ablation}
\begin{figure}[t]
\centering
\includegraphics[width=0.65\linewidth]{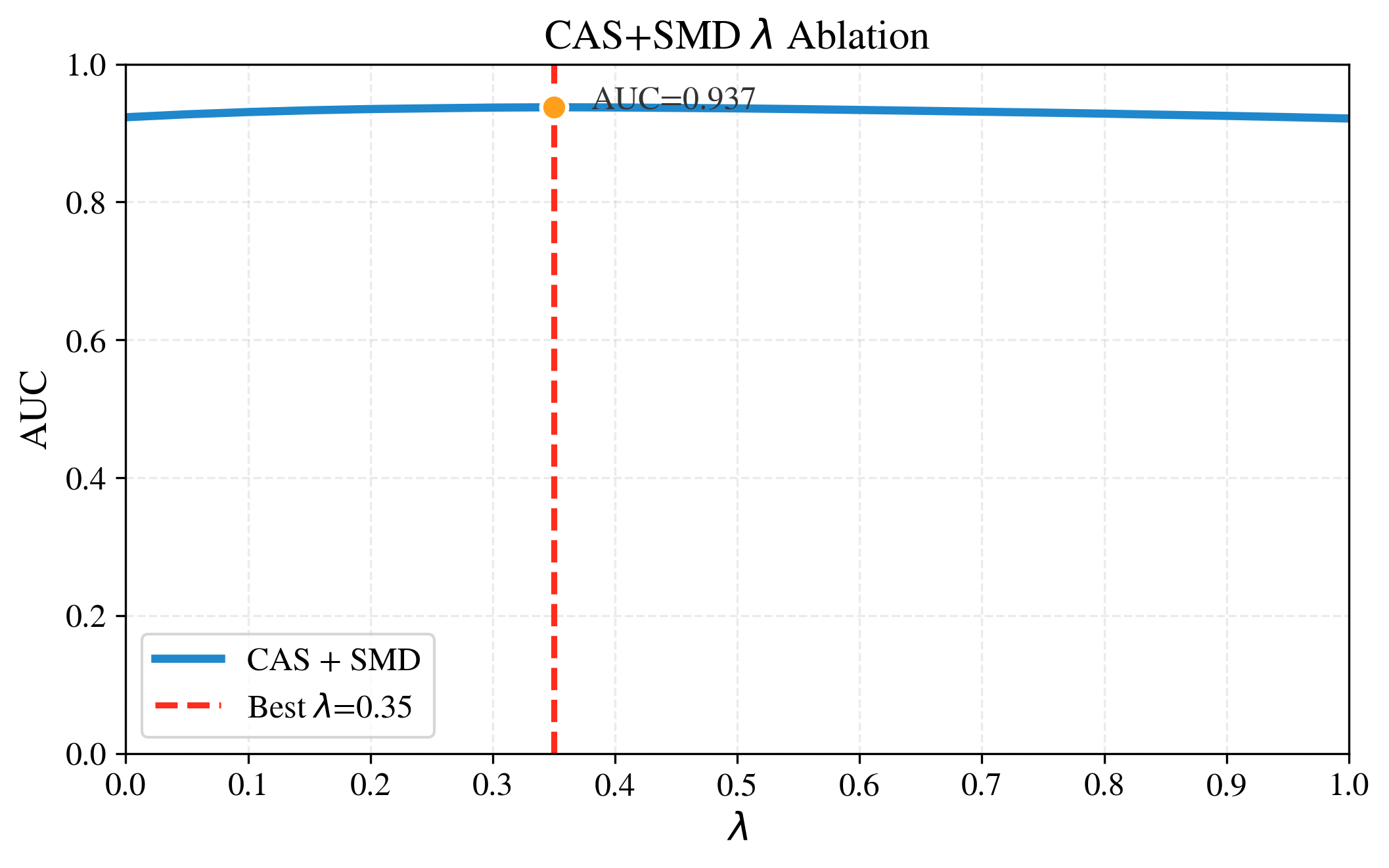}
\setlength{\abovecaptionskip}{-2pt}
\caption{The AUC sensitivity of the combined CAS+SMD score as a function of
$\lambda$. Performance remains stable across a broad range of weights. The best
result occurs at $\lambda=0.35$, indicating that the combination of CAS and SMD
is not sensitive to a narrow choice of the weighting parameter.}
\label{fig:lambda_auc_ablation}
\end{figure}

For the combined reliability score, we first $z$-normalize the rollout-level
CAS and SMD scores:
\begin{equation}
    \tilde U = \frac{U-\mu_U}{\sigma_U+\epsilon},
\end{equation}
where $\mu_U$ and $\sigma_U$ are the mean and standard deviation of the
corresponding score vector and $\epsilon$ is a small numerical constant. We
then compute
\begin{equation}
    U_{\mathrm{CAS+SMD}} = \lambda \tilde{U}_a + (1-\lambda)\tilde{U}_s,
\end{equation}
where $\tilde{U}_a$ and $\tilde{U}_s$ are the normalized CAS and SMD scores,
respectively. The main reliability table uses $\lambda=0.35$. To evaluate the
sensitivity of this choice, we sweep $\lambda$ from 0 to 1 and report AUC on
the validation/test rollouts for success-versus-failure prediction. This
analysis checks whether the combined score remains competitive across a range
of weights and whether the selected weight improves over using CAS or SMD
alone.  Fig.~\ref{fig:lambda_auc_ablation} shows the resulting sensitivity
curve.

\subsection{Real-World Deployment Results}
\label{app:real_robot}
The real-robot evaluation uses six UR5 spatial manipulation tasks: (i) move the
wipes bottle near the red cup; (ii) move the red container near the blue bin;
(iii) move the tissue box farther from the orange cup; (iv) move the yellow box
to the empty space between the two cubes; (v) pick up the gray cube and place
it on the white plate; and (vi) pick up the red cup and place it in the black
container. These tasks vary object size, spatial relation, and workspace
configuration, and are used to test whether the same action-selection and
monitoring signals remain useful on physical hardware.

\subsubsection{Success-Failure Separation on Real-Robot Rollouts}
\label{app:real_auc}
\begin{table}[t]
\centering
\begin{tabular}{lc}
\toprule
\textbf{Uncertainty Metric} & \textbf{AUC} \\
\midrule
TB-TP & 0.661 \\
TB-PCS & 0.662 \\
TB-E & 0.705 \\
TB-D & 0.713 \\
A-PI & 0.719 \\
A-VI & 0.746 \\
A-AI & 0.742 \\
\midrule
CAS & \underline{0.759} \\
SMD & \textbf{0.912} \\
\bottomrule
\end{tabular}
\setlength{\abovecaptionskip}{3pt}
\caption{AUC scores for predicting failure on real-robot trajectories.}
\label{tab:real-robot_auc_scores}
\end{table}

\begin{figure*}[t]
\centering
\setlength{\belowcaptionskip}{12pt}
\begin{minipage}{0.24\textwidth}\centering\includegraphics[width=\linewidth]{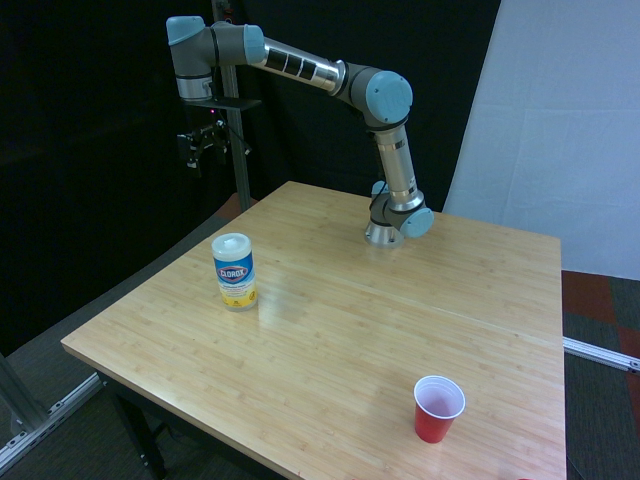}\end{minipage}\hfill
\begin{minipage}{0.24\textwidth}\centering\includegraphics[width=\linewidth]{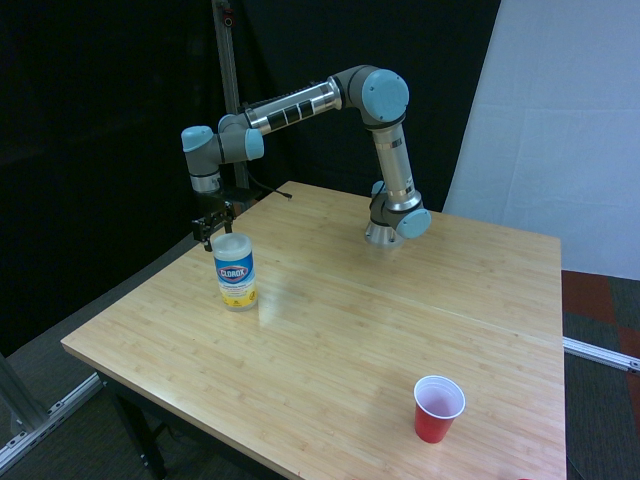}\end{minipage}\hfill
\begin{minipage}{0.24\textwidth}\centering\includegraphics[width=\linewidth]{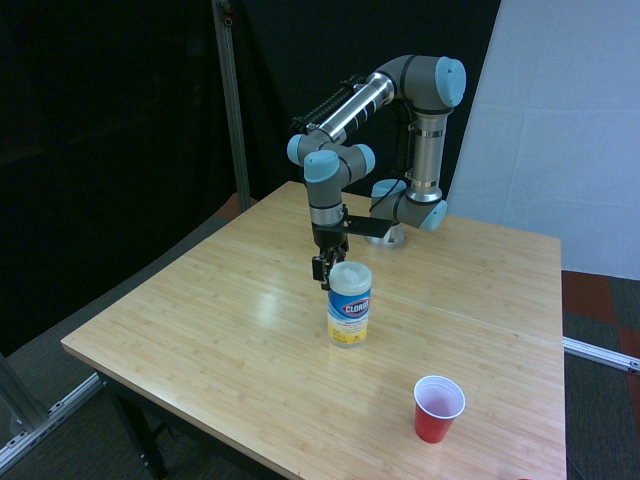}\end{minipage}\hfill
\begin{minipage}{0.24\textwidth}\centering\includegraphics[width=\linewidth]{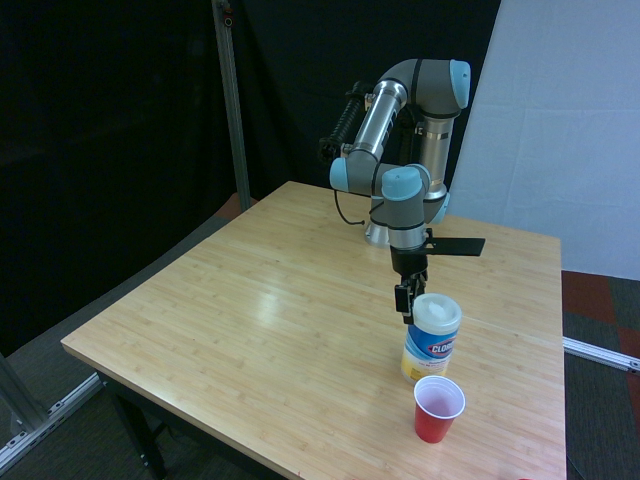}\end{minipage}
\vspace{2mm}
\begin{minipage}{0.24\textwidth}\centering\includegraphics[width=\linewidth]{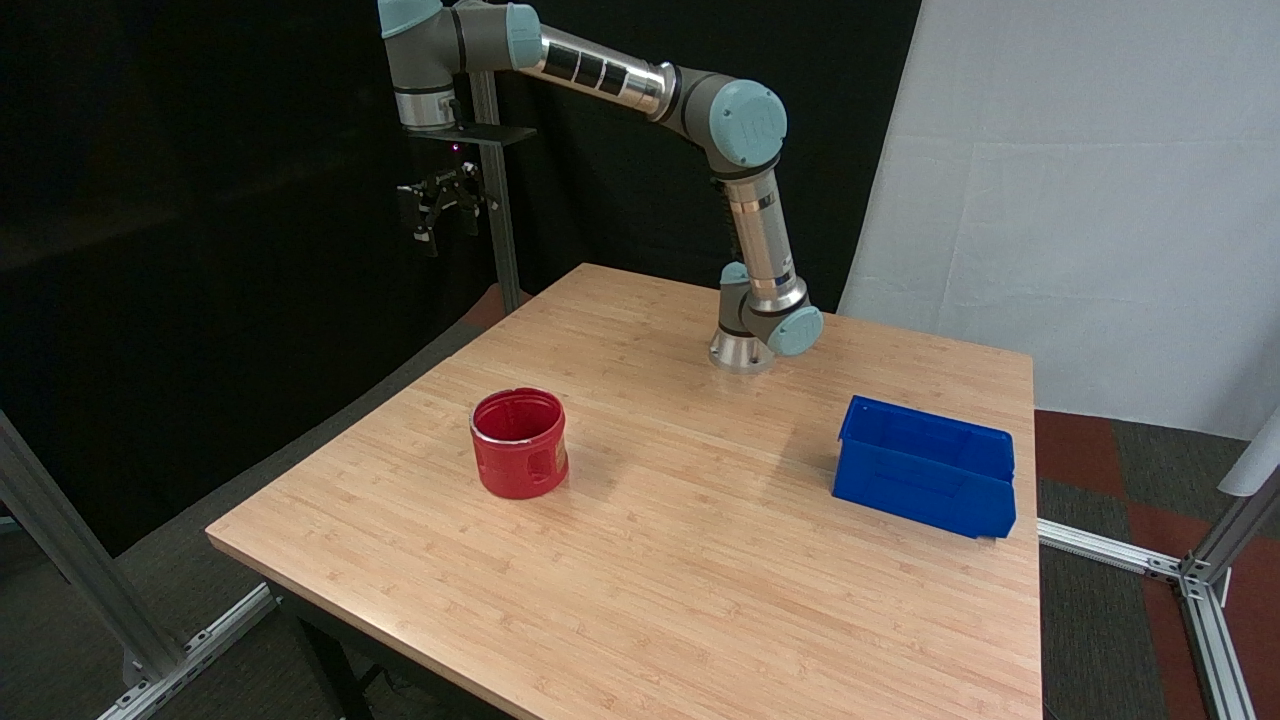}\end{minipage}\hfill
\begin{minipage}{0.24\textwidth}\centering\includegraphics[width=\linewidth]{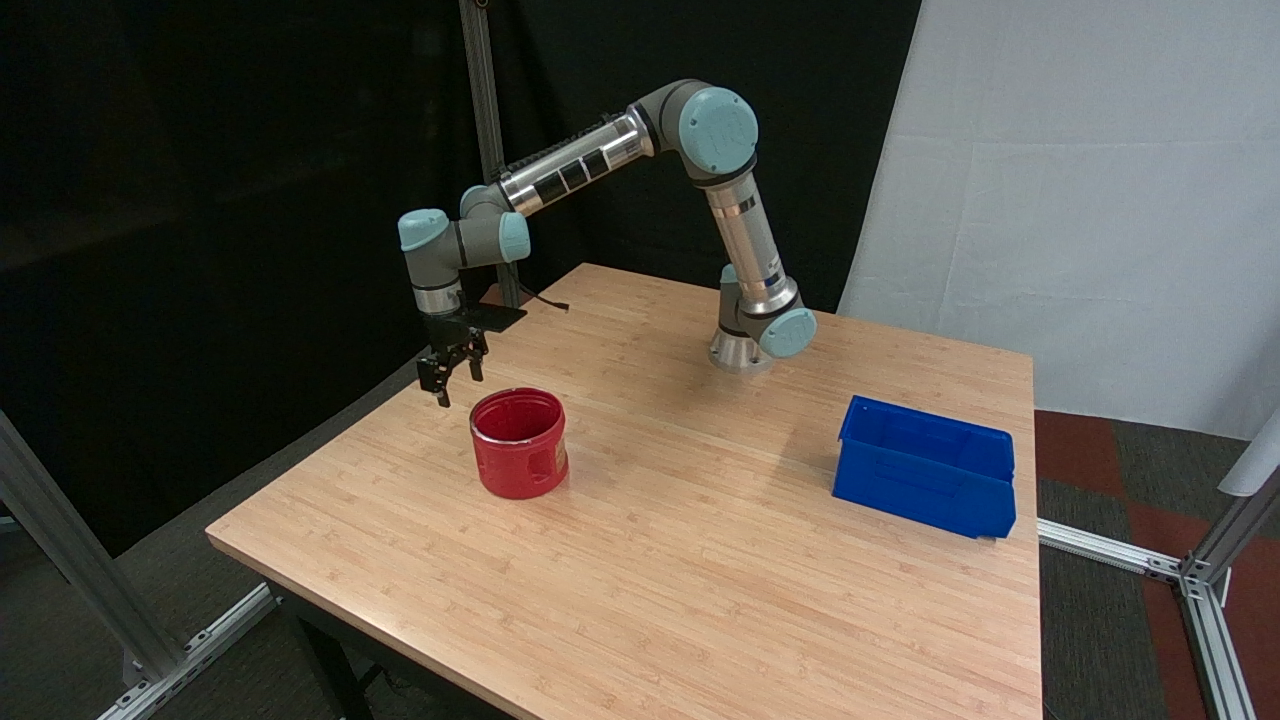}\end{minipage}\hfill
\begin{minipage}{0.24\textwidth}\centering\includegraphics[width=\linewidth]{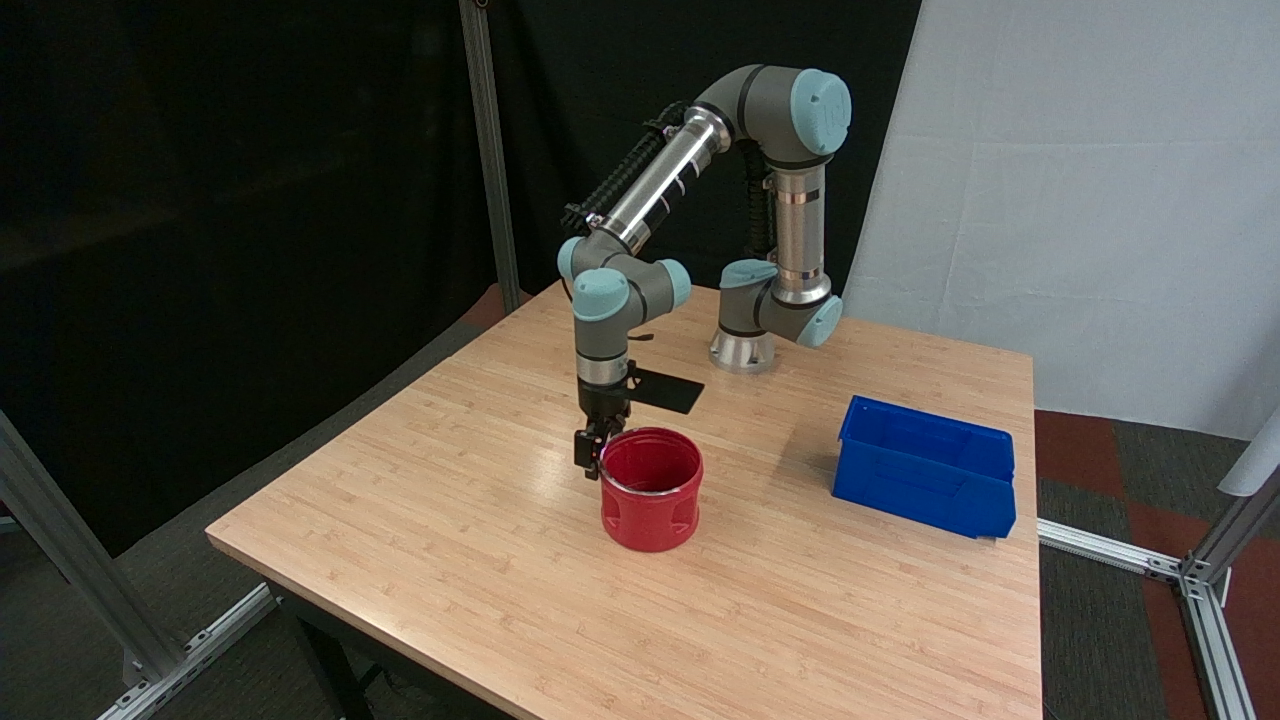}\end{minipage}\hfill
\begin{minipage}{0.24\textwidth}\centering\includegraphics[width=\linewidth]{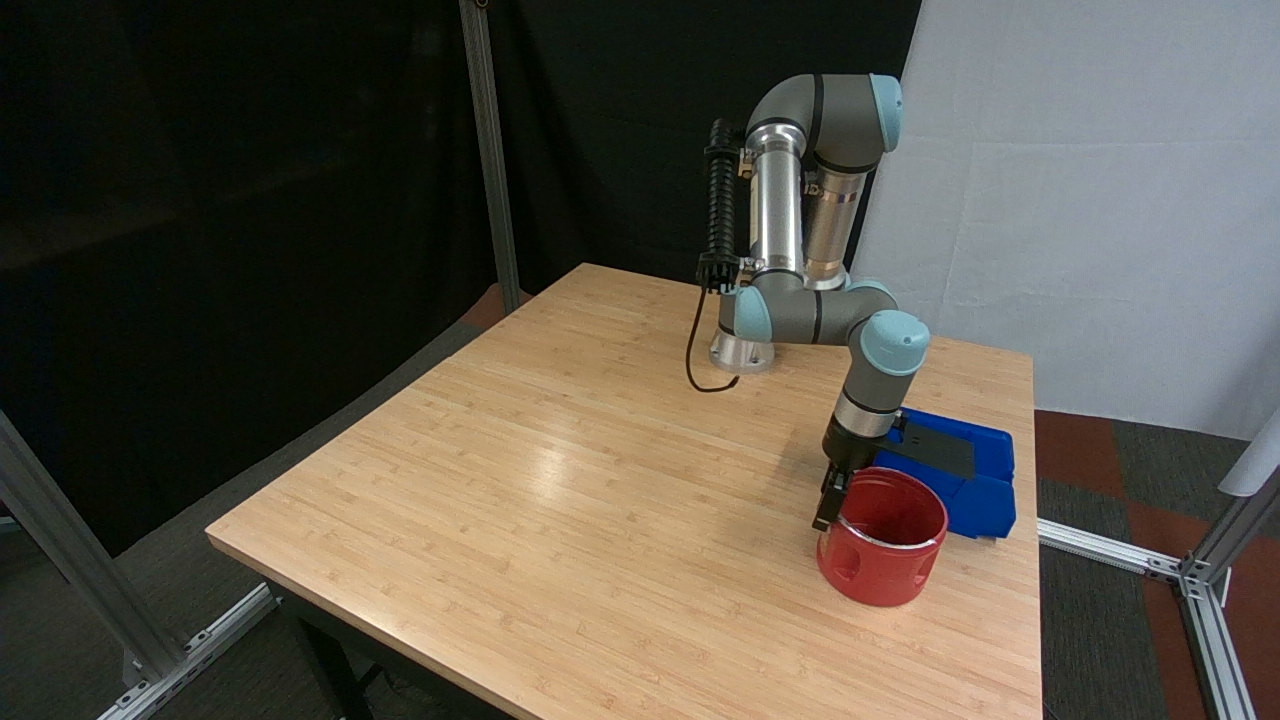}\end{minipage}
\vspace{2mm}
\begin{minipage}{0.24\textwidth}\centering\includegraphics[width=\linewidth]{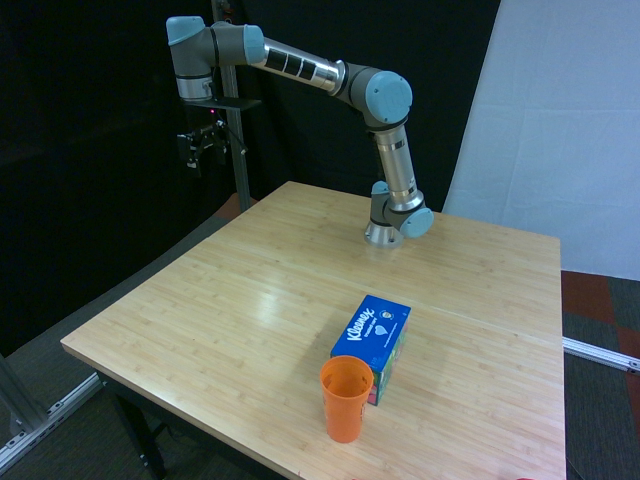}\end{minipage}\hfill
\begin{minipage}{0.24\textwidth}\centering\includegraphics[width=\linewidth]{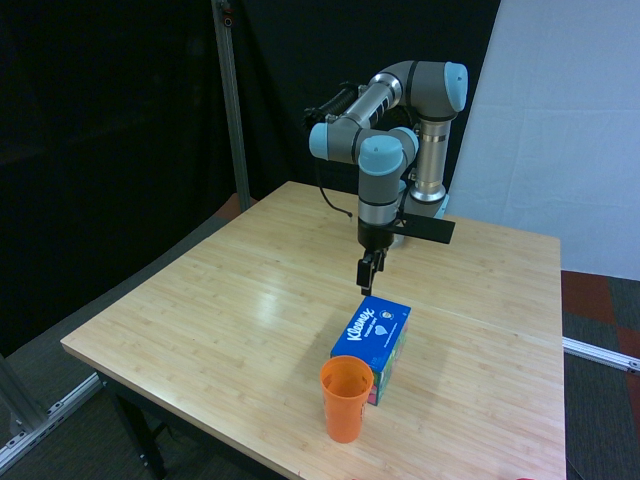}\end{minipage}\hfill
\begin{minipage}{0.24\textwidth}\centering\includegraphics[width=\linewidth]{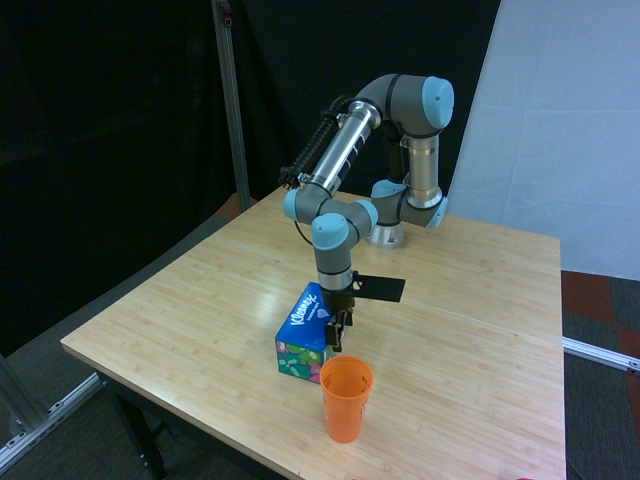}\end{minipage}\hfill
\begin{minipage}{0.24\textwidth}\centering\includegraphics[width=\linewidth]{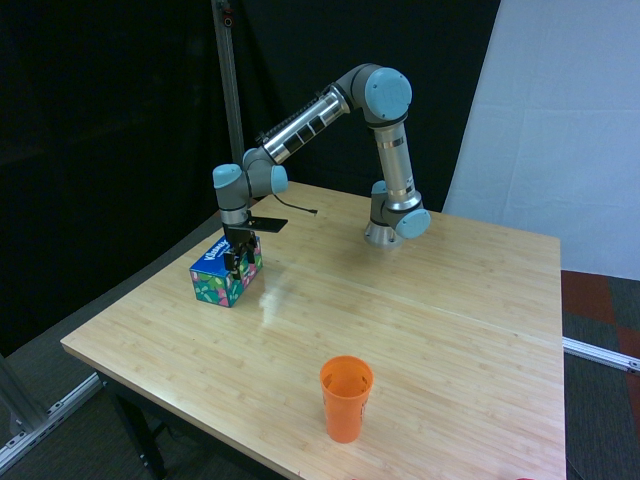}\end{minipage}
\vspace{2mm}
\begin{minipage}{0.24\textwidth}\centering\includegraphics[width=\linewidth]{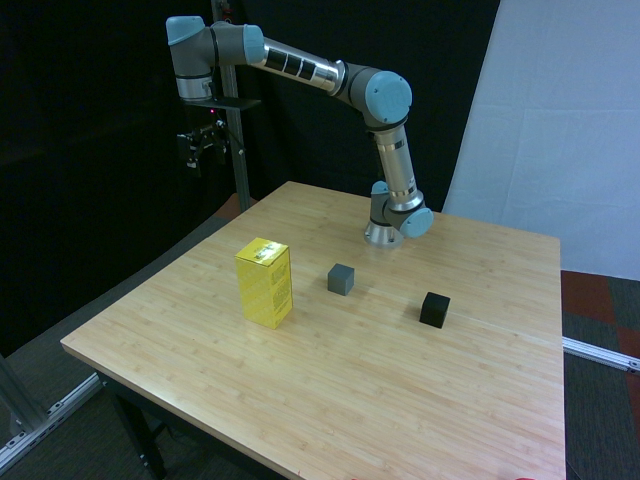}\end{minipage}\hfill
\begin{minipage}{0.24\textwidth}\centering\includegraphics[width=\linewidth]{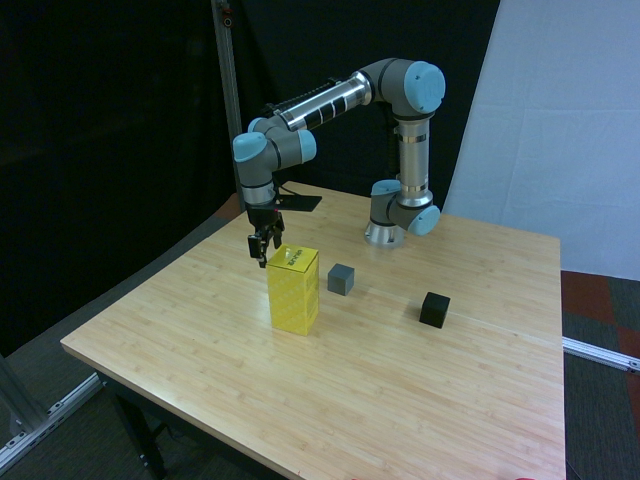}\end{minipage}\hfill
\begin{minipage}{0.24\textwidth}\centering\includegraphics[width=\linewidth]{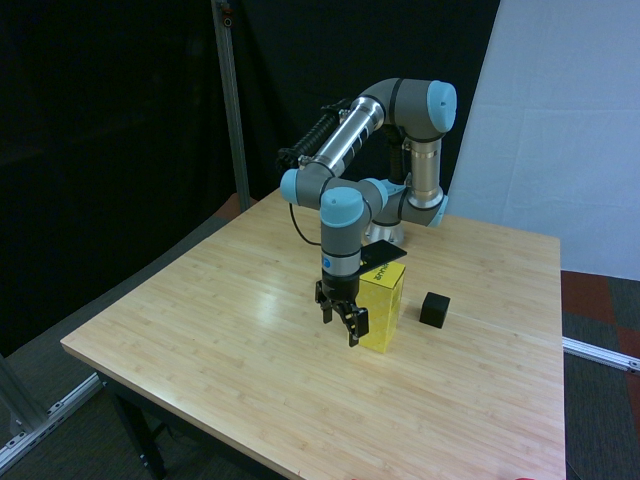}\end{minipage}\hfill
\begin{minipage}{0.24\textwidth}\centering\includegraphics[width=\linewidth]{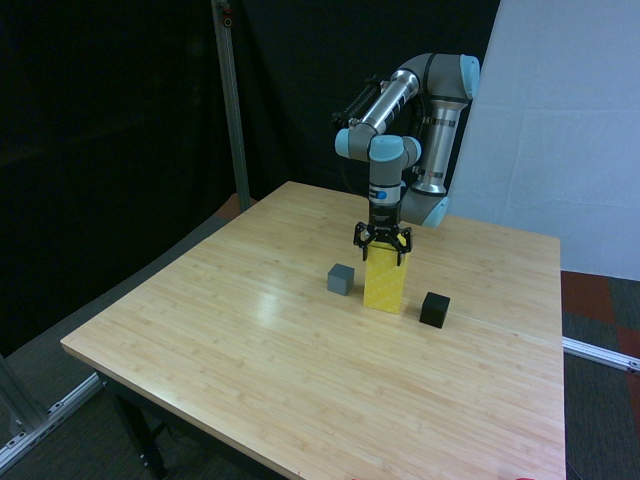}\end{minipage}
\vspace{2mm}
\begin{minipage}{0.24\textwidth}\centering\includegraphics[width=\linewidth]{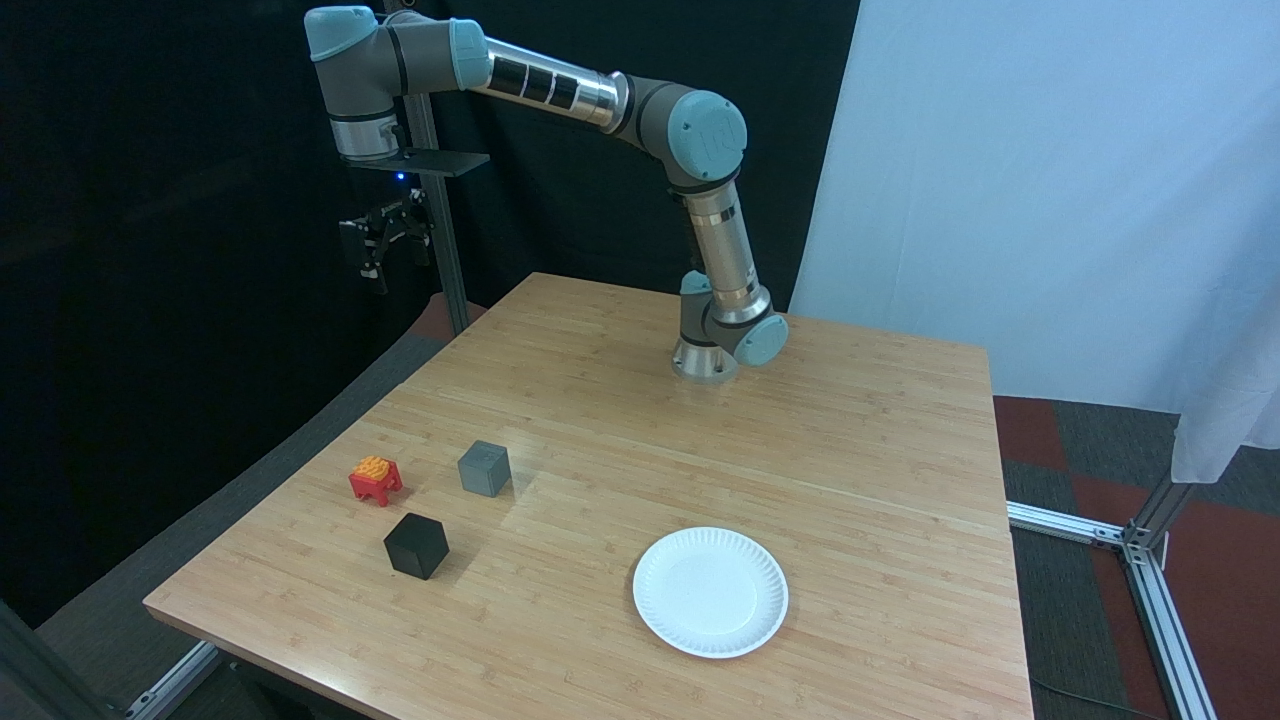}\end{minipage}\hfill
\begin{minipage}{0.24\textwidth}\centering\includegraphics[width=\linewidth]{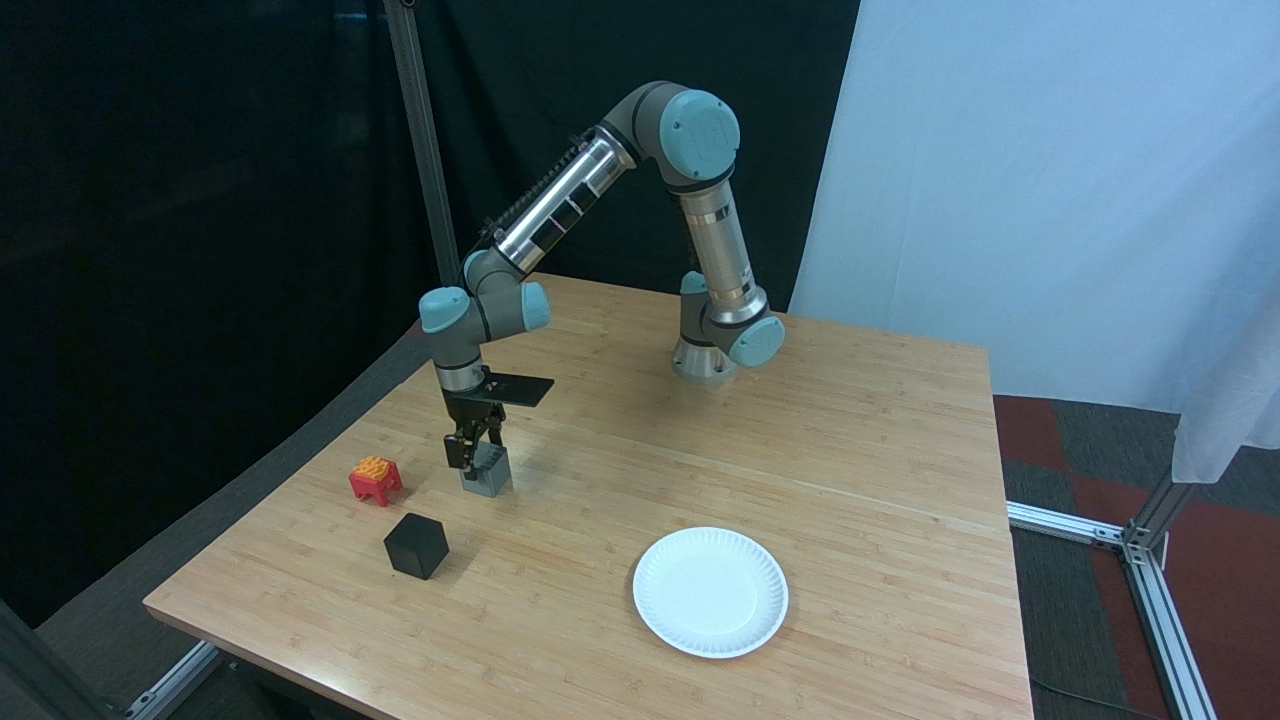}\end{minipage}\hfill
\begin{minipage}{0.24\textwidth}\centering\includegraphics[width=\linewidth]{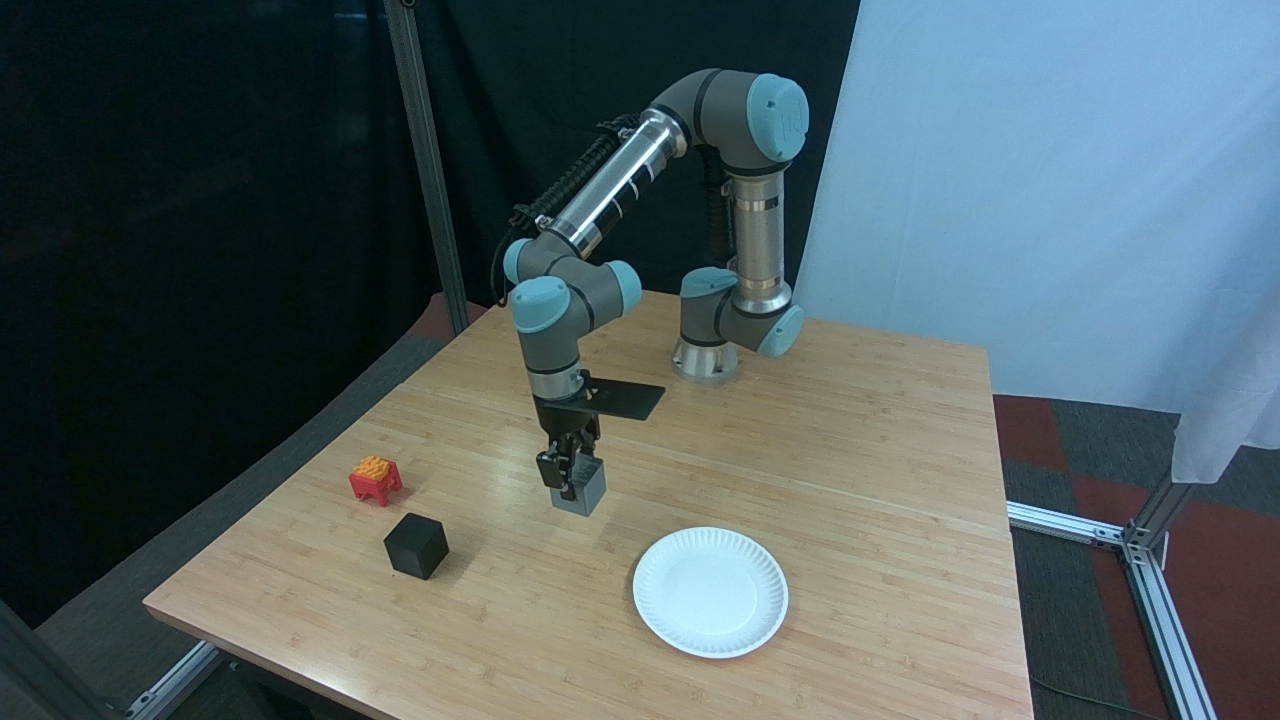}\end{minipage}\hfill
\begin{minipage}{0.24\textwidth}\centering\includegraphics[width=\linewidth]{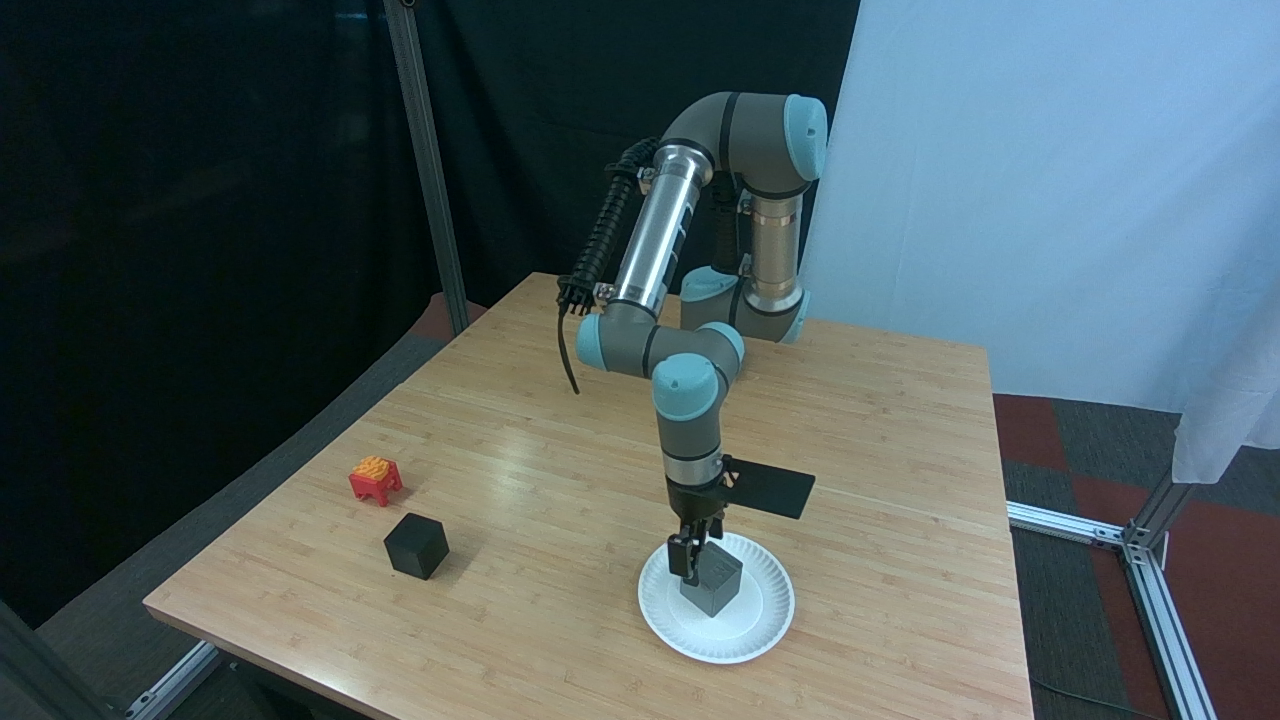}\end{minipage}
\vspace{2mm}
\begin{minipage}{0.24\textwidth}\centering\includegraphics[width=\linewidth]{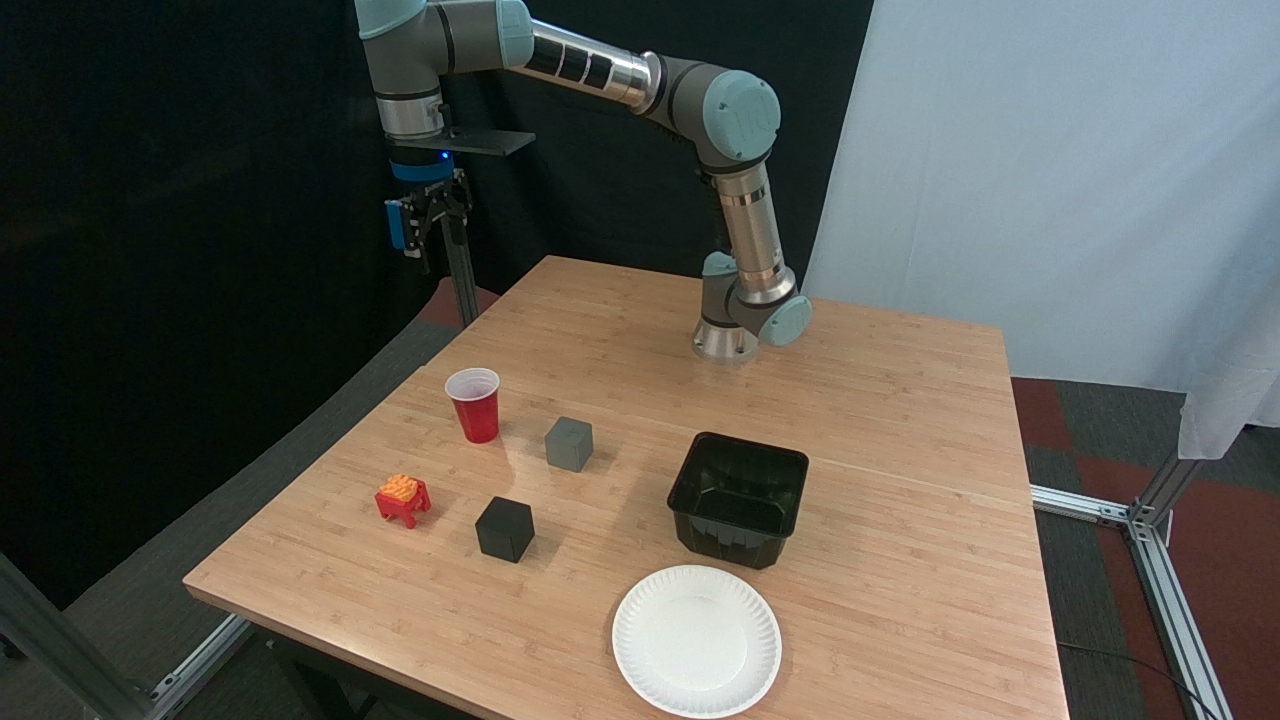}\end{minipage}\hfill
\begin{minipage}{0.24\textwidth}\centering\includegraphics[width=\linewidth]{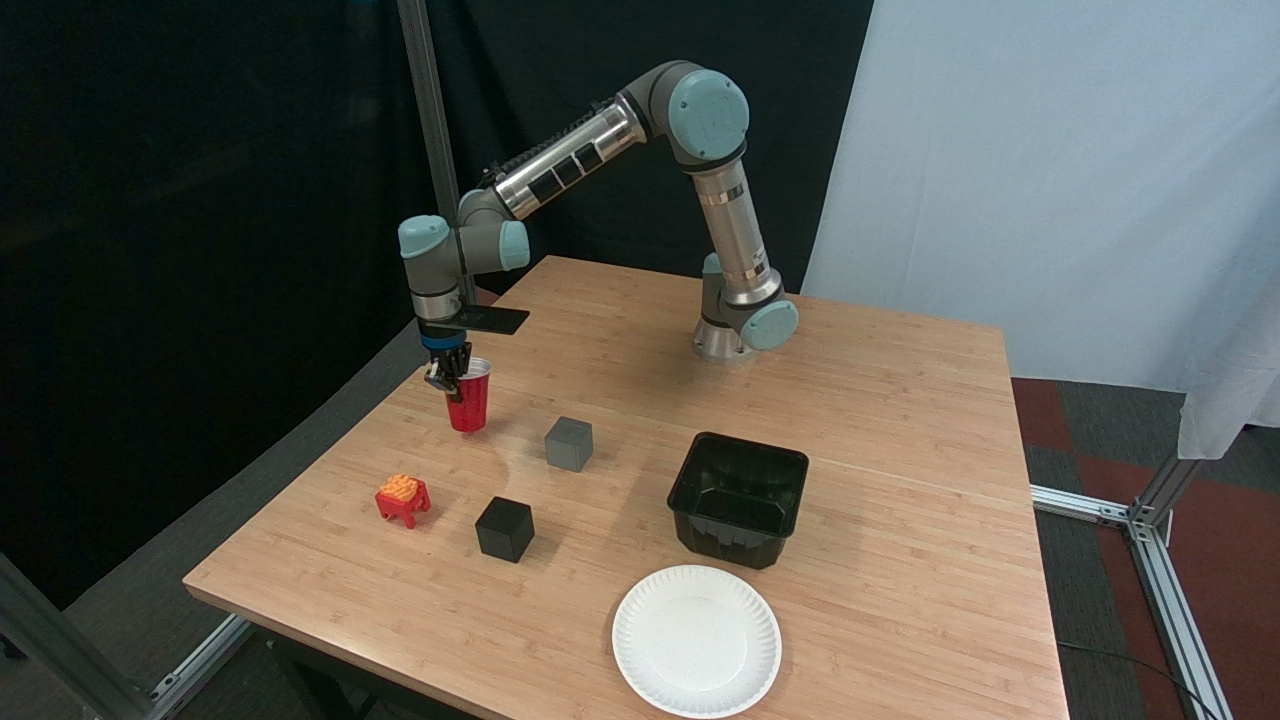}\end{minipage}\hfill
\begin{minipage}{0.24\textwidth}\centering\includegraphics[width=\linewidth]{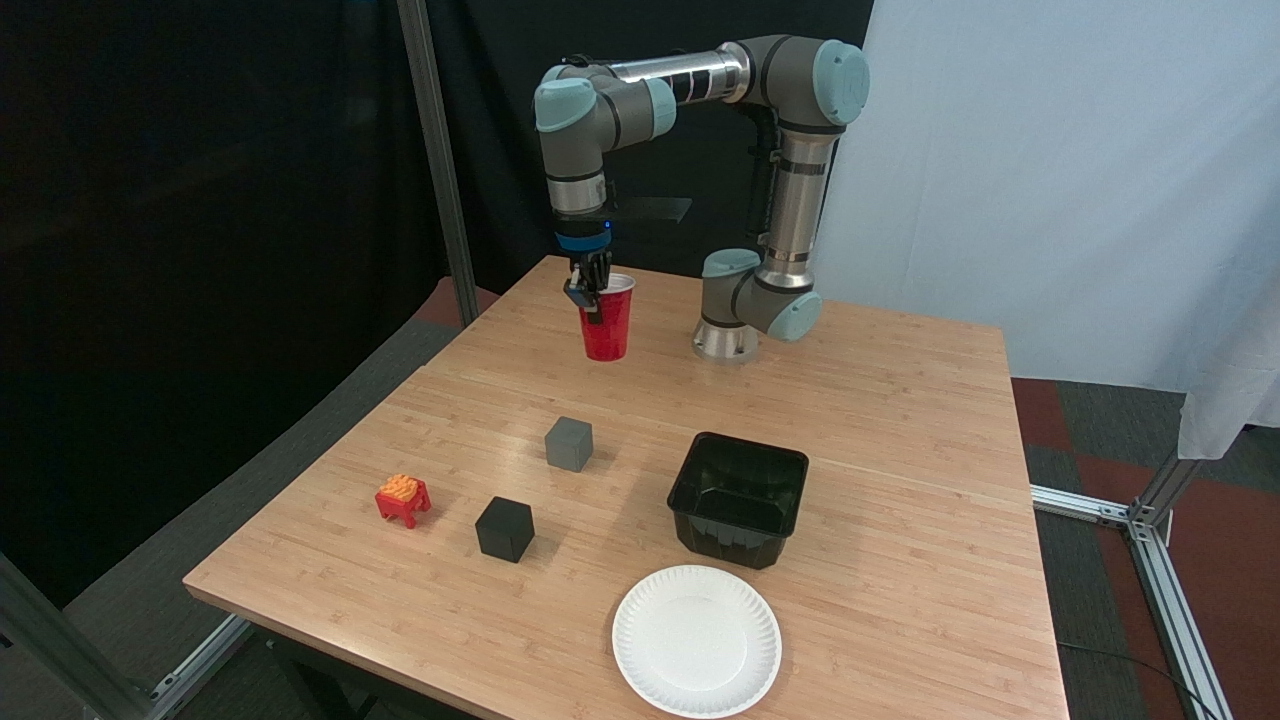}\end{minipage}\hfill
\begin{minipage}{0.24\textwidth}\centering\includegraphics[width=\linewidth]{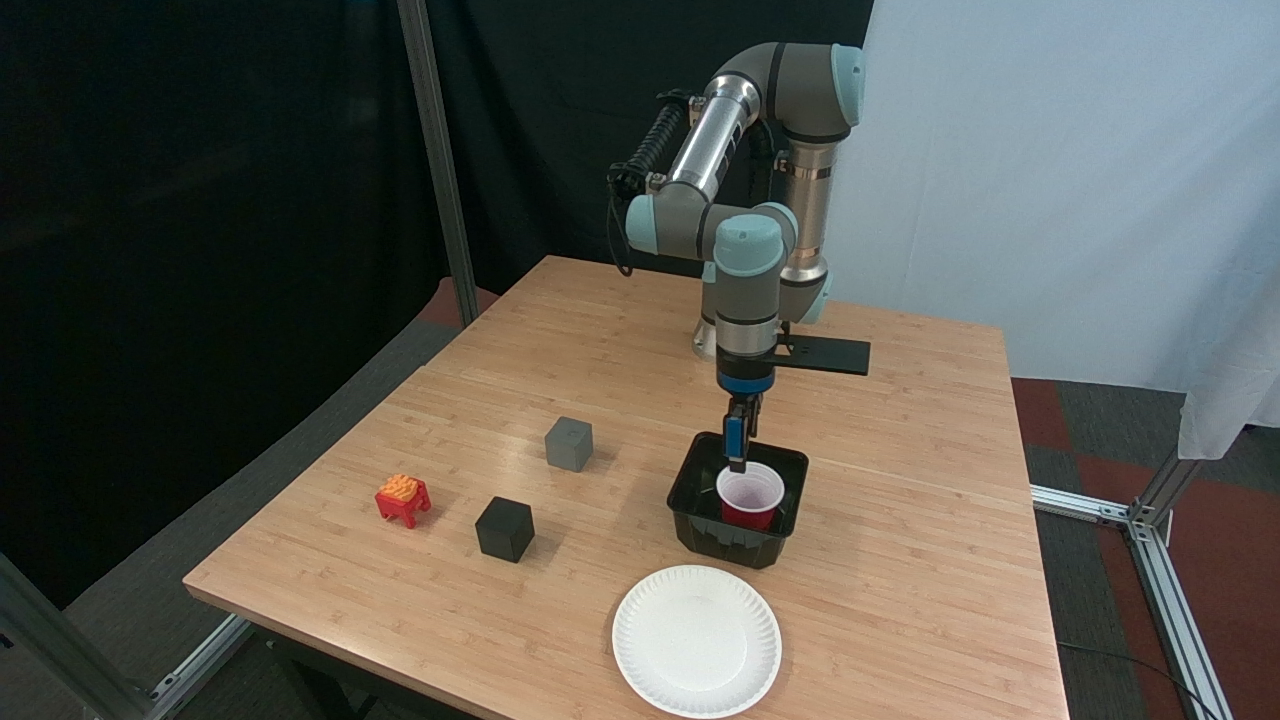}\end{minipage}
\caption{Example execution frames for the six real-robot spatial manipulation
tasks: (top row) \textit{Move the wipes bottle near the red cup}; (second row)
\textit{Move the red container near the blue bin}; (third row) \textit{Move the
tissue box farther from the orange cup}; (fourth row) \textit{Move the yellow
box to the empty space between the two cubes}; (fifth row) \textit{Pick up the
gray cube and place it on the white plate}; (bottom row) \textit{Pick up the
red cup and place it in the black container}.}
\label{fig:robot_demonstrations}
\end{figure*}

For each task, we collect 20 successful and 20 failed trajectories. Each
trajectory is generated by the deployed $\pi_0$ policy under the nominal scene
configuration.  For each rollout, we compute the seven EU-VLA metrics along
with CAS and SMD. Table~\ref{tab:real-robot_auc_scores} reports the AUC scores
for success-failure prediction. SMD gives the strongest hardware reliability
signal, and CAS is the strongest action-level score among the evaluated
methods, indicating that the proposed signals transfer to physical hardware.
Fig.~\ref{fig:robot_demonstrations} shows representative rollouts for the six
tasks.

\subsubsection{Action-Level Reliability Under Distribution Shifts}
\label{app:real_shift}
We evaluate action-level performance under three policies: default $\pi_0$,
mean-action baseline with 10 noisy samples averaged, and CAS. We measure
success rates on Task 1 and Task 5.  Each task is executed under four scene
conditions: standard, lighting shift, camera perturbation, and cluttered scene.
For every combination of task, condition, and policy, we run 30 trials.
\begin{table}[t]
\centering
\begin{tabular}{llccc}
\toprule
\textbf{Task} & \textbf{Condition} & \textbf{Default} $\boldsymbol{\pi_0}$ & \textbf{Mean} & \textbf{CAS} \\
\midrule
\multirow{4}{*}{Task 1}
 & Standard & 0.50 & 0.60 & \textbf{0.70} \\
 & Lighting shift & 0.47 & 0.57 & \textbf{0.70} \\
 & Camera perturbation & 0.43 & 0.57 & \textbf{0.67} \\
 & Cluttered scene & 0.43 & 0.53 & \textbf{0.63} \\
\midrule
& Average & 0.46 & 0.57 & \textbf{0.68} \\
\midrule
\multirow{4}{*}{Task 5}
 & Standard & 0.47 & 0.60 & \textbf{0.70} \\
 & Lighting shift & 0.43 & 0.57 & \textbf{0.67} \\
 & Camera perturbation & 0.43 & 0.53 & \textbf{0.67} \\
 & Cluttered scene & 0.40 & 0.50 & \textbf{0.63} \\
\midrule
& Average & 0.43 & 0.55 & \textbf{0.67} \\
\bottomrule
\end{tabular}
\setlength{\abovecaptionskip}{3pt}
\caption{Real-robot success rates under mild distribution shifts.}
\label{tab:ur5_success_rate}
\end{table}
Table~\ref{tab:ur5_success_rate} summarizes the real-robot success rates.
Across both tasks and all evaluated perturbations, CAS maintains higher success
rates than the default $\pi_0$ policy and the mean-action baseline, showing
that the action-selection strategy remains useful under these mild scene
changes.

\subsubsection{State-Level Failure Detection at the Workspace Edge}
\label{app:workspace_edge}
Lastly, we evaluate SMD in a controlled failure setting where the robot
approaches the edge of its reachable workspace. Since deliberately executing
catastrophic failures such as tipping or collision is infeasible on hardware,
we place the target object beyond the reachable region so that the rollout
naturally pushes the arm toward its joint limits. We use Task 1 and run 20
trials for each policy: default $\pi_0$ without SMD gating, and $\pi_0$ with
SMD gating.

We define three regions within the workspace: a safe region inside the nominal
task space, a warning zone where the robot approaches joint limits or hardware
constraints, and a failure region where the robot hits its limits, enters
protect mode, or hard-stops. In default $\pi_0$ rollouts, the robot continues
until the UR5 built-in safety system triggers a protective stop. With SMD
gating, execution pauses when the latent-state anomaly score exceeds the
threshold, allowing the robot to wait safely for intervention without entering
protect mode.

\begin{table}[t]
\centering
\begin{tabular}{lcc}
\toprule
 & $\boldsymbol{\pi_0}$ \textbf{(No SMD)} & \textbf{With SMD Gate} \\
\midrule
Protective-stop events & 20 / 20 & \textbf{4 / 20} \\
Proactive SMD halts & -- & 16 / 20 \\
\bottomrule
\end{tabular}
\setlength{\abovecaptionskip}{3pt}
\caption{State-level failure detection at the workspace edge.}
\label{tab:real_robot_failure_detection}
\end{table}

\begin{figure}[t]
\centering
\includegraphics[width=1\linewidth]{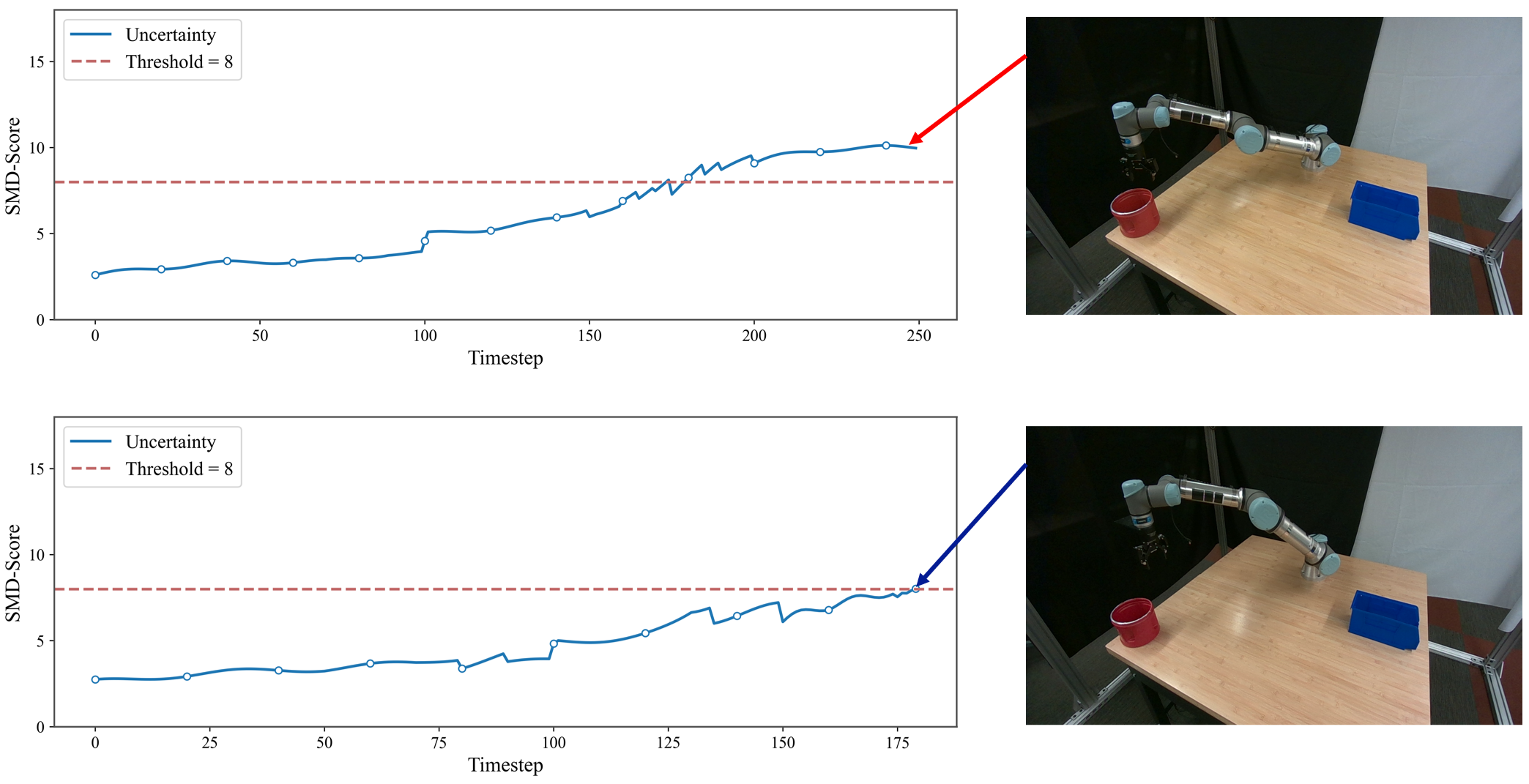}
\caption{An illustration of SMD behavior in real-robot workspace-edge trials.
Top: under the default $\pi_0$ policy without SMD, the robot continues
executing until it reaches hardware limits and triggers a protective stop.
Bottom: with SMD gating, the rising SMD score crosses the threshold earlier in
the trajectory, causing the robot to halt proactively before entering the
failure region.}
\label{fig:proactive_vs_protective}
\end{figure}

Table~\ref{tab:real_robot_failure_detection} reports the aggregated statistics,
and Fig.~\ref{fig:proactive_vs_protective} illustrates representative
workspace-edge rollouts. SMD gating reduces protective stops from 20/20 to 4/20
and proactively halts 16/20 trajectories before a protective stop occurs.
Across both action-selection and failure-detection settings, CAS reduces
high-variance action errors and SMD provides early warnings of abnormal
execution contexts, supporting safer VLA-based control on the robot arm.

\end{document}